**Title:** Enhancing drug and cell line representations via contrastive learning for improved

anti-cancer drug prioritization

**Authors:**

Patrick J. Lawrence[1]: patrick.skillman-lawrence@osumc.edu

Xia Ning Ph.D.[1,2,3,*]: ning.104@osu.edu

**Affiliations:**
1) Biomedical Informatics Department, The Ohio State University, 1800 Cannon Drive, Lincoln Tower 250, Columbus, OH 43210, USA.
2) Computer Science and Engineering Department, The Ohio State University, 2015 Neil Avenue, Columbus, OH 43210, USA.
3) Translational Data Analytics Institute, The Ohio State University, 1760 Neil Avenue, Columbus, OH 43210, USA.
*) Corresponding Author

**Abstract:** Due to cancer's complex nature and variable response to therapy, precision oncology informed by omics sequence analysis has become the current standard of care. However, the amount of data produced for each patients makes it difficult to quickly identify the best treatment regimen. Moreover, limited data availability has hindered computational methods' abilities to learn patterns associated with effective drug-cell line pairs. In this work, we propose the use of contrastive learning to improve learned drug and cell line representations by preserving relationship structures associated with drug mechanism of action and cell line cancer types. In addition to achieving enhanced performance relative to a state-of-the-art method, we find that classifiers using our learned representations exhibit a more balances reliance on drug- and cell line-derived features when making predictions. This facilitates more personalized drug prioritizations that are informed by signals related to drug resistance.

**Introduction:** Cancer, the leading cause of death worldwide, remains a challenge to treat due to its complex nature and variable response to therapy, even among patients with the same cancer type. Omics, which describes the collective and comprehensive analysis of biomolecular data, is being leveraged to conduct precision oncology. Transcriptomics, a subfield of omics that studies gene expression by quantifying relative levels of RNA molecules, has been an especially valuable tool. Oncologists use transcriptomics—comparing normal and tumor cells—to identify changes in gene expression. These alterations are used to pinpoint the molecular process(es) driving tumorigenesis in each patient, allowing clinicians to develop personalized treatment recommendations. However, because RNA sequencing (RNA-seq) measures the expression of more than 20,000 genes, manually evaluating each patient's data to determine the best treatment is neither scalable nor pragmatic.

Instead, machine learning models have been applied to omics data to predict cellular sensitivities to drug candidates. RefDNN[1], one state-of-the-art (SOTA) method, represents drugs via their structural similarity to a set of reference drugs. It then leverages these reference drugs to produce cell line representations: each dimension is the output of an Elastic Net[2] model trained on transcriptomic data to predict a cell line's sensitivity to a distinct reference drug. During evaluation, RefDNN applies a dense neural network (DNN) to the Hadamard product of drug and cell line representations to predict cancer drug response (CDR). However, RefDNN is limited by data quality: it requires complete CDR data for all reference drug and cell line pairs during training.

DeepDSC[3], another SOTA method, also leverages transcriptomic data to produce cell line representations. It does so via the latent embeddings of a pretrained autoencoder (AE). By encoding cell lines' transcriptomic profiles into a lower-dimensional space, the AE captures key information while mitigating the risk of overfitting, which commonly occurs when training deep learning models on limited data. In DeepDSC, drugs are represented by Morgan fingerprints[4]. A dense neural network (DNN) is then applied to drug-cell line pairs to predict CDR. DeepDSC is more robust than RefDNN as it can leverage incomplete data during training. However, its use of generic fingerprints may still hinder performance as they are not customized to predict CDR. Similarly, the cell line features derived from DeepDSC's AE are optimized for transcriptomic profile reconstruction, meaning they, too, may not be relevant for CDR prediction.

As such, two major challenges in achieving precise CDR predictions are the creation of problem-specific embeddings and the integration of drug-cell line pairs. Both of which are complicated by the quality and availability of data. To address these issues, we propose a novel framework, denoted as SiamCDR, that leverages contrastive loss within a **Siam**ese neural network[5] (SNN) to enhance the expressiveness of drug and cell line representations, thereby improving **C**ancer **D**rug **R**esponse predictions (**Figure 1a**). Specifically, our model learns to project drugs and cell lines to embedding spaces that encode the similarities of gene targets for drugs and cancer type for cell lines, respectively. This is guided by the intuition that drugs with similar targets will have similar effects. Additionally, drug efficacies among cells of the same cancer type should be more similar than drug efficacies among cells of different cancers. The benefit of using SNNs here is their, and other few shot learning frameworks', ability to learn such similarity relationships from a limited number of training instances for each group[5–7]. Specifically, drugs are grouped by their mechanism of action (MOA); cell lines are grouped by their cancer type. Moreover, contrastive learning will ensure our method preserves similarity relationships tailored to predicting CDR.

Our experiments show that SiamCDR produces higher-quality and more personalized drug prioritizations than DeepDSC. In fact, a network analysis of genes whose expression is significantly correlated with SiamCDR's docetaxel prioritization in breast cancer identified enriched pathways known to induce docetaxel-resistance. This suggests the SiamCDR can learn to modulate its recommendations from transcriptomic signals associated with drug efficacy. Finally, using SiamCDR, we identify multiple repurposing candidates for difficult-to-treat cancers.

**Results:**

<u>Evaluating model performance</u>: High performing models will prioritize the most effective drugs for each cell line above ineffective drugs. To that end, we assess model performance using average $P_{\text{cell}}@k$ ($k = 1, 2, 3, 4, 5, 10$) and average $P_{\text{cancer}}@k$ ($k = 1, 2, 3, 4, 5$) defined by Equations (2) and (3), respectively. These measure the average proportion of highly effective drugs among a model's top-$k$ prioritized drugs. We compare the performance of our top-3 model architectures SiamCDR_LR, SiamCDR_RF, and SiamCDR_DNN against the DeepDSC baseline (see Methods section for details about each model). **Tables 1a** and **1b** report the average $P_{\text{cell}}@k$ for cell lines with trained-on and novel cancers, respectively; **Tables 2a** and **2b** report the average $P_{\text{cancer}}@k$ for cell lines with trained-on and novel cancers, respectively. **Tables 1** and **2** also report Bonferroni-corrected significance levels with respect to differences in performance relative to DeepDSC. Each SiamCDR model achieves improvements over DeepDSC for all reported metrics in **Tables 1** and **2**. However, the Bonferroni-corrected significance level of these improvements varies. Note, we do not observe significant differences in the performance of the three SiamCDR models. This indicates all three are equally proficient at recommending effective drugs to target the cells of trained-on and novel cancers.

In **Table 1a**, we observe no significant difference in performance between the models for $P_{\text{cell}}@1$. However, for all other values of $k$, we observe significant improvements in performance for all three SiamCDR models compared to DeepDSC. Notably, the significance level increases with $k$ for all SiamCDR models. This indicates that, for trained-on cancers, SiamCDR models prioritize a greater number of effective drugs at the very top compared to DeepDSC, which suggests that SiamCDR is more robust than DeepDSC with respect to quality of its prioritizations. Identical trends are observed when performance is generalized to cancer type ($P_{\mathrm{cancer}}$) in **Table 2a**, apart from SiamCDRRF gaining significance at $k = 1$.

**Table1b** reports significant improvements in $P_{\mathrm{cell}}@1$ compared to DeepDSC for all SiamCDR models. However, SiamCDRLR loses its significance for $k = 2$, while both SiamCDRRF and SiamCDRDNN do not achieve significant improvement for $k = 3$. When generalized to cancer type ($P_{\mathrm{cancer}}$) in **Table 2b**, we observe substantial jumps in performance for DeepDSC, SiamCDRLR, and SiamCDRDNN. Conversely, SiamCDRRF demonstrates more stable performance, which combined with the improved relative performance of DeepDSC, yields a loss of significance at $k = 4$. This suggests DeepDSC, SiamCDRLR, and SiamCDRDNN all may be more sensitive to cell line outliers than SiamCDRRF. One potential explanation is these models have learned to identify broad-spectrum (or cancer-specific) anti-cancer drugs. In which case, presenting cell lines with resistances to these drugs would yield reduced performance. However, averaging $P_{\mathrm{cell}}$ by cancer would reduce the impact each individual outliers has, thereby raising performance. If this were the case, that would indicate that SiamCDRRF has greater capacity for producing tailored cell line predictions compared to the other models.

To test whether DeepDSC, SiamCDRLR, and SiamCDRDNN models may, in fact, favor broad-spectrum anti-cancer drugs more compared to SiamCDRRF, we visualize prediction trends for each drug-cell line pair. In **Figure 2**, we plot continuous effective scores (y-axis)—denoted as CES and defined by Equation (1)—against CDRs predicted by **a**) DeepDSC, **b**) SiamCDR$_{RF}$, **c**) SiamCDR$_{LR}$, and **d**) SiamCDR$_{DNN}$ (x-axis). The binarization threshold for CDR labels is illustrated by horizontal, dashed lines. Pairs above this line are highly effective (top-10% with respect to CDR). To specifically assess drugs commonly predicted as effective, we filter the drugs to include only those prioritized among the top-5 for at least three cell lines by all four models. Each of the six identified drugs are highlighted with distinct shapes and colors in **Figure 2**. Note that to qualitatively examine the relationship between predicted score and CES, the scale of the x-axis in **Figure 2a** has been adjusted; the range of predictions produced by DeepDSC (0.068 to 0.259) is much smaller than the range of the SiamCDR models' predictions (0.0 to 1.0).

We observe a positive association between CES and model predictions for each model. Considering only the six highlighted drugs, DeepDSC and SiamCDR$_{RF}$ (**Figures 2a** and **b**) produce scores with large drug-wise variance, suggesting their predictions are more informed by cell lines than either SiamCDR$_{LR}$ or SiamCDR$_{DNN}$. Conversely, in **Figures 2c** and **d**, we observe that SiamCDR$_{LR}$ and SiamCDR$_{DNN}$'s predictions have small by-drug variance. This indicates that both have learned to identify broad-spectrum anti-cancer drugs. With respect to SiamCDR$_{LR}$, it is likely that the relationship(s) that determine cell lines' individual drug responses are too complex to capture via logistic regression. Additionally, it is probable that there is insufficient data to train all SiamCDR$_{DNN}$'s parameters. As a result, both models have learned to identify broad-spectrum anti-cancer drugs as a way of minimizing loss during training. SiamCDR$_{RF}$'s ensemble approach allows decision trees to capture non-linear interactions more effectively than LRs while being less prone to overfitting than DNNs, allowing it to provide unique predictions for each drug-cell line pair. This is supported by the average variance in SiamCDR$_{LR}$'s, SiamCDR$_{RF}$'s, and SiamCDR$_{DNN}$'s predicted scores for any drug with a predicted score above 0.5 for at least one cell line. SiamCDR$_{RF}$ achieves 5.11- and 2.72-times greater variance in its predictions than SiamCDR$_{LR}$ and SiamCDR$_{DNN}$, respectively, indicating it may be a more suitable candidate for precision medicine applications.

Identifying drug- and cell line-derived feature importance to model predictions. We measure model feature importance (see Methods) to discern whether SiamCDR$_{RF}$'s tailored drug prioritizations result from a more balanced influence of drug- and cell line-derived features compared to DeepDSC, SiamCDR$_{LR}$ or SiamCDR$_{DNN}$. In **Figure 3**, we rank features by their importance to the predictions of **a)** DeepDSC, **b)** SiamCDR$_{RF,}$ **c)** SiamCDR$_{LR,}$ or **d)** SiamCDR$_{DNN}$. For each subplot, a maximum of the top-100, non-zero features are depicted. Relative feature importance is conveyed by bar height, and feature source—drug or cell line—is represented by color. The average importance for each source is denoted by a horizontal line of the source's respective color.

In **Figure 3a**, we observe that DeepDSC's predictions are more heavily influenced by cell lines than drugs with 90% of the top-10 features being cell line-derived. Conversely, we find SiamCDR$_{LR}$ and SiamCDR$_{DNN}$ almost exclusively prefer drug-derived features (**Figures 3c** and **d**, respectively). None of SiamCDR$_{LR}$'s top-100 features are derived from cell lines and SiamCDR$_{DNN}$ preferred a single cell line-derived feature among its top-100 (46[th]). The lack of cell line influence yields non-personalized drug prioritization and explains the small by-drug variance in SiamCDR$_{LR}$'s and SiamCDR$_{DNN}$'s predictions observed in **Figure 2c** and **d**.

On the other hand, in **Figure 2b**, we observe a balanced influence of both drugs and cell lines on SiamCDR$_{RF}$'s predictions: among the top-10 features, four and six are drug- and cell line-derived, respectively. In addition, we observe that the average relative feature importance for drugs and for cell lines is more similar in magnitude for SiamCDR$_{RF}$ than for any of the other models. This balanced integration of drug and cell line information when making predictions highlights SiamCDR$_{RF}$'s practical utility over both SiamCDR$_{LR}$ and SiamCDR$_{DNN}$ for precision medicine. Therefore, we consider only SiamCDR$_{RF}$ for the remaining experiments.

Comparing the expressiveness of DeepDSC's and SiamCDR$_{RF}$'s cell line representations. Both DeepDSC and SiamCDR$_{RF}$ conduct representation learning to produce cell line embeddings ($\mathbf{e}_{ae}$ and $\mathbf{e}_c$, respectively). However, $\mathbf{e}_c$ is learned via contrastive loss within an SNN framework, making it more likely to have captured task-specific information than $\mathbf{e}_{ae}$. This is important as increasing the expressiveness of embeddings will enhance a model's ability to distinguish cell lines of different cancer types. The autoencoder framework that produces $\mathbf{e}_{ae}$ has been optimized for reconstruction, making it less likely to have captured the optimal information to predict CDR. To evaluate whether using $\mathbf{e}_c$ may better segregate cell lines by cancer than $\mathbf{e}_{ae}$, we visualize the clustering of cell lines represented by DeepDSC's $\mathbf{e}_{ae}$ (**Figure 4a**) and SiamCDR$_{RF}$'s $\mathbf{e}_c$ (**Figure 4b**) with t-SNE plots. Only cancer types with more than 15 cell lines are shown.

We also calculate $\mathbf{e}_c$ and $\mathbf{e}_{ae}$'s intra-group similarity and inter-group separability, where groups are defined as cancers. Use of $\mathbf{e}_c$ reduces intra-cancer similarity by $-8.85\%$ compared to $\mathbf{e}_{ae}$, with mean inter-group similarities of 0.91 and 1.00. This indicates each cancer's cell lines are not as tightly clustered when using $\mathbf{e}_c$ versus $\mathbf{e}_{ae}$. Interestingly, a mean intra-group similarity of 1.00 suggests $\mathbf{e}_{ae}$ may not be able to differentiate between cell lines of the same cancer and may consider them identical. In terms of inter-cancer type separability, $\mathbf{e}_c$ achieves a significant improvement of 77.03% compared to $\mathbf{e}_{ae}$ (1.77 vs 1.00; $p < 0.0001$). Importantly, a separability score of 1.0 conveys that cancers cannot be differentiated when using $\mathbf{e}_{ae}$. The inability of $\mathbf{e}_{ae}$ to discern cells of the same and different cancer(s) is supported by the lack of well-defined clusters in **Figure 4a**.

Conversely, $\mathbf{e}_c$'s inter-cancer type separability implies that it can better distinguish between cell lines of distinct cancers. This is supported by **Figure 4b**, which illustrates that SiamCDR$_{RF}$'s $\mathbf{e}_c$ separates into distinct clustering structures. For example, all cell lines of lung cancer, leukemia, lymphoma, myeloma, and skin cancer, respectively, exist in single, well-defined clusters for their respective cancer type (circled with dotted lines). The distance of these clusters relative to others corresponds to a limited overlap in tumorigenesis for these cancers. Other cancer types, including bone cancer, sarcoma, brain cancer, and neuroblastoma, exist within well-defined clusters that are geometrically close to other cancer clusters. This is highlighted in *cluster i* in **Figure 4b** (magnified in **Figure 4c**). The proximity of brain cancer and neuroblastoma is unsurprising given both involve tumorigenesis within the nervous system. Likewise, the proximity of bone cancer and sarcoma is expected as sarcomas often originate in bone tissue. Previous evidence also demonstrates a close relationship between some forms of sarcoma and neuroblastomas[8]. *Cluster ii*, magnified in **Figure 4d**, represents another multi-cancer grouping (breast cancer, rhabdoid, and kidney). The close proximity of rhabdoid tumors and kidney cancers in the embedding space corresponds to the fact that rhabdoid tuners often originate in the kidneys, indicating that the embeddings accurately capture cancer relations[9]. Additionally, the proximity between kidney and breast cancers may indicate that embedding geometry may even capture a cell line's driving mutation(s) (e.g., the risk for both breast and kidney cancer is significantly elevated by *PTEN* mutations[10]). Finally, we observe in *cluster iii* of **Figure 4b** (magnified in **Figure 4e**) that the embeddings of colorectal, gastric, bile, and pancreatic cancer cell lines are in close geometric proximity to one another, which may loosely represent gastrointestinal cancers. Each set of multi-cancer groupings further demonstrates the capacity of SiamCDR's framework to embed highly nuanced relationships, thereby producing embeddings ($\mathbf{e}_c$) with high expressiveness that are more fine-tuned to predict CDR than DeepDSC's embeddings ($\mathbf{e}_{ae}$).

Comparing the expressiveness of SiamCDR$_{RF}$'s and DeepDSC's drug representations. The drug representations ($\mathbf{e}_d$) produced by SiamCDR's framework should also capture more task-specific information than the Morgan fingerprints ($\mathbf{f}$) employed by DeepDSC. This is because the fingerprints are generic, sparse vectors produced from predefined heuristics. To assess the expressiveness of each representation scheme, we use t-SNE plots to visualize how drugs are clustered when using either $\mathbf{f}$ (**Figure 4f**) or $\mathbf{e}_d$ (**Figure 4g**). Only MOAs with more than 10 drugs present in the pretraining data are highlighted. The included MOAs and their drug counts are reported in **Supplementary Table 8**.

We also calculate intra-MOA similarity and inter-MOA separability for both **f** and $\mathbf{e}_d$, finding that the use of $\mathbf{e}_d$ to significantly improves intra-MOA similarity and inter-MOA separability compared to **f** by 89.50% (0.77 and 0.40) and 60.82% (1.92 and 1.19), respectively. The significance level for each is less than 0.0001. Interestingly, this improvement demonstrates that preserving drug relations based on shared gene target captures rich semantic information related to MOA. Moreover, the improvement in both metrics garnered by $\mathbf{e}_d$ indicates $\mathbf{e}_d$ has higher expressiveness than **f**.

This is further illustrated by the t-SNE plots. We observe, in **Figure 4g**, multiple well-defined clusters corresponding to distinct MOAs. This contrasts with largely unsegregated MOAs in **Figure 4f**. Interestingly, in **Figure 4g,** *cluster iv* (magnified in **Figure 4h**) is comprised of both phosphatidylinositol-3-kinase (PI3K) inhibitors and mammalian target of rapamycin (mTOR) inhibitors and is highly separated from all other MOAs. Both PI3K and mTOR belong to a signaling pathway that controls cell growth and survival[11]. Because genes comprising these pathways may be closely linked, the clustering of drugs by pathways may result from overlapping genetic targets of drugs spanning these MOA.

Additionally, some MOAs, such as protein synthesis inhibitors (PSI), do not segregate well and span multiple clusters. Protein synthesis represents a broad category modulated by many genes, which like genetic pathways, may overlap with other MOAs. For example, in *cluster v* of **Figure 4g**, we observe three PSIs near tubulin polymerization inhibitors (TPIs). C*luster v* is magnified in **Figure 4i**.  One such PSI, brefeldin-a (BFA), targets *ARF1,* which controls protein secretion and coordinates tubulin polymerization[12,13]. Thus, it is not unexpected to observe BFA closely associated with TPIs. In another example, CUDC-907 (fimepinostat), classified as a PI3K inhibitor, is found among HDAC inhibitors in *cluster vi* (magnified in **Figure 4j**). HDACs facilitate histone modification, modulating gene accessibility for transcription to control gene expression[14]. This reclassification is supported by recent evidence investigating its dual-inhibitory properties[15]. The work concluded that CUDC-907's HDAC inhibition elicits more significant changes to gene expression than the those driven by its PI3K inhibition. As such, the drug embeddings produced by SiamCDR$_{RF}$ may improve MOA classifications, a conclusion supported by the classification of embeddings produced by SiamCDR$_{RF}$ for novel drugs. We embed two drugs, YM-201636 and BNC-105, which are missing gene target data and were excluded from pretraining but have reported MOAs (PI3K inhibitor and TPI). We find that both embeddings place the drugs within the clusters of their respective MOAs (*clusters v* and *vi*, respectively). Altogether, this evidence supports the use of our proposed framework for learning biologically relevant drug representations.

Evaluating FDA-approved drug prioritizations. Because prioritization quality can be difficult to ascertain when evaluating novel candidates, we first compare SiamCDR$_{RF}$'s and DeepDSC's ability to prioritize FDA-approved drugs. If a model can highly prioritize known/approved treatments, it is likely that other highly prioritized candidates may also be effective. Specifically, we examine how highly each model prioritizes the subset of drugs FDA-approved to treat each cell line based on its cancer type compared to all 369 drugs with at least one reported indication. We leverage 805 unseen cell lines for which there was also at least one FDA-approved drug among our data. We score all combinations of these drugs and cell lines using both DeepDSC and SiamCDR$_{RF}$. For each cell line, both its top-prioritized FDA-approved therapy and the average priority for all of its FDA-approved drug(s) are presented in **Supplementary Table 9**. In **Table 3**, we report by-cancer summaries of these results and include significance levels comparing SiamCDR$_{RF}$'s and DeepDSC's mean and max prioritizations of FDA-approved treatments; SiamCDR$_{RF}$ achieves significant improvements in mean and max prioritization in 40 and 35% of the evaluated cancers, respectively. Despite not achieving improvements across the majority of evaluated cancers, SiamCDR$_{RF}$'s prioritizes a unique FDA-approved drug top in 80% of cancers. Conversely, DeepDSC exhibits limited uniqueness among its top-prioritized drugs; for 50% of the evaluated cancers either docetaxel or doxorubicin is DeepDSC's top-prioritized FDA-approved therapy. This implies DeepDSC has learned to prioritize a few broad-spectrum anti-cancer drugs rather personalized candidates, which may be driven by an inability to discern individual cells after they have been embedded as illustrated in **Figure 4a**. This notion is further supported by the standard deviation (std) in the priority of each cancer's top-prioritized FDA-approved therapies for each model. The mean std of SiamCDR$_{RF}$'s 24 top-prioritized drugs is 17.32, while for DeepDSC's 13 drugs it is 1.36. This indicates that while DeepDSC's predictions do vary by cell line (**Figure 2a**), the priority-order of drugs does not change. The variance in by-cell line variance in score may indicate DeepDSC has learned to identify cell lines that have fewer effective drugs. However, the lack of individualized drug prioritization suggests that, like SiamCDR$_{DNN}$, DeepDSC may have overfit to the data.

Case Study 1: Exploring docetaxel prioritization in breast cancer: To better understand what SiamCDR$_{RF}$ may be leveraging to tailor its prioritizations, we explore SiamCDR$_{RF}$'s prioritization of docetaxel for breast cancer. This pair was chosen as it follows a bimodal distribution—the model prioritizes docetaxel very highly for some breast cancers and lowly for others. We measure the correlation in expression for each of the 463 genes with SiamCDR$_{RF}$'s prioritization of docetaxel across all breast cancer cells to ascertain whether transcriptomic differences in breast cancer cells have influenced docetaxel's priority. In **Supplementary Table 10**, we report 40 genes whose expression is significantly correlated (magnitude above 0.35 and significance below 0.1). We use STRING[16] to perform network analysis on the isolated genes and observe significant enrichment of the PI3K-Akt (19 genes), MAPK (11 genes), and RAS (12 genes) signaling pathways. This aligns with documented evidence of transcriptomic changes in *Akt* pathways inducing docetaxel resistance[17–20]. Moreover, this suggests SiamCDR$_{RF}$ has learned to leverage signals associated with transcriptomic modulation of CDR to tailor its predictions, making it a valuable tool for precision oncology.

Case study 2: Examining repurposing candidates for difficult to treat cancers. Lastly, after demonstrating high-performance and personalized prioritizations, we evaluate SiamCDR$_{RF}$'s capacity to repurposing candidates for cancer treatment. We consider four cancers: two *trained-on* (bladder, BC; and head and neck, HNC) and two *novel* (gastric, GC; and prostate, PC). These were selected via **Table 3** from the subset of cancers with fewer than 10 approved therapies and an average max priority for prioritized FDA-approved therapies less than 30. Because SiamCDR$_{RF}$ consistently prioritizes FDA-approved therapies highly, the other high priority candidates may also be effective. Moreover, the limited number of approved therapies imply theses cancers are difficult to treat.

For each of these cancers' cell lines, we obtain SiamCDR$_{RF}$'s prioritizations across all 1,119 drugs. Then, for each cancer, we examine the literature for anti-cancer evidence associated with any drug prioritized more highly, on average, than the top-prioritized FDA-approved drug recorded in **Table 3**. We present the drug candidates with positive evidence in **Table 4**. MOAs, gene targets, and indications are obtained from the Broad's drug repurposing hub (DRH)[21]. The first row for each cancer in **Table 4** is its top-prioritized FDA-approved drug. For cancers whose highest prioritized FDA-approved drug is above 50, only candidates prioritized among the top-50 are considered.

For BC, we found 13/45 (28.9%) drugs prioritized above its top-prioritized FDA-approved drug, valrubicin, with anti-BC evidence. The mean priority of the anti-BC candidates was 18.8 compared to valrubicin's mean priority of 70.6. Notably, dolastatin-10 (dol-10) (mean priority: 5.0) was approved to treat BC after the LINCS and DRH data were published. In addition, there are several drugs with published *in vivo* anti-BC evidence. Cabazitaxel was found to increase objective response in muscle invasive bladder cancer by over 2 times the current gold standard (26% to 57%) during a phase II clinical trial[22]. Oltipraz, reduced bladder carcinogenesis by detoxifying a bladder-specific carcinogen in mice[23]. In another mouse model, GZD824 displayed activity against FGFR1-mutant BCs, which are especially difficult to treat[24]. Gemcitabine and 10-hydroxycamptothecin (HCPT), two drugs with demonstrated anti-BC activity, both became more effective when administered in combination with triptolide[25,26]. Similarly, SiamCDR$_{RF}$ also highly prioritized triptolide to treat both GC and PC: *in vitro*, triptolide reportedly enhanced apoptotic activity of other anti-cancer drugs when used to pretreat GC cells[27] and inhibited PC cell growth[28].

For HNC, 8/44 (18.2%) with higher priority than the highest prioritized FDA-approved drug, docetaxel, have anti-HNC evidence. The mean priority of these drugs was 31.9 compared to docetaxel's priority of 85.1. In clinical trials, poziotinib, litronesib, ninlaro, and temsirolimus each positively affected HNC progression[29–32]. Additionally, previously published *in vitro* and *in vivo* studies present evidence of anti-HNC activity for YM-155, BGT226, and SN-38[33–35].

For GC, 17.8% of drugs (8/45) with higher priority than the highest prioritized FDA-approved drug, docetaxel, have evidence of anti-GC activity (mean priority of 12.8 and 79.4, respectively). Specifically, alvespimycin, cabazitaxel, and exatecan-mesylate (ExM) each achieved positive results in clinical trials[36–38]. Furthermore, three additional drugs, including triptolide, exhibited positive, pre-clinical evidence: romidepsin and YM-155[39,40]. Another highly prioritized drug, BGT226, lacks published evidence exploring its anti-GC activity; however, one of its gene targets, *PIK3CA*, is a common oncogene known to stimulate GC tumorigenesis. As such, further investigation into this candidate is warranted.

Finally, 43.8% of the drugs (8/16) with higher priority than cabazitaxel, PC's top-prioritized FDA-approved drug have documented evidence of anti-BC activity. The mean priority of these drugs is 8.4 compared to cabazitaxel's mean priority of 17.7. Notably, YM-155 was recently approved to treat PC. Alvespimycin demonstrated capacity to achieve complete responses in patients with PC during phase I clinical trials[38]. One *in vitro* study cited ExM as the most potent TOP-INH against PC cells. HCPT, dol-10, and camptothecin each showcased positive pre-clinical evidence as well[41–43].

**Discussion:**

Due to variable drug responses among patients with the same cancer type, optimizing cancer treatment for each patient remains a challenge. Computational methods have been proposed to predict CDR, but their performance is limited by data availability and modeling strategies. To address this, we propose SiamCDR$_{RF}$, which uses SNNs to pretrain drug and cell line encoders to produce embeddings with high expressiveness. SNNs excel in scenarios with limited data availability as they focus on learning the (dis)similarity between training instances, which enables fine-grained differences, relevant to predicting CDR, to be captured. SiamCDR$_{RF}$ uses a RF model to predict CDR from the learned embeddings. Its balance of simplicity, to prevent overfitting, with an ability to capture complex, non-linear relationships, affords the model enhanced precision in its pan-cancer drug prioritizations over LR and DNN classifiers.

Furthermore, we find SiamCDR$_{RF}$ achieves significant improvements in performance compared to the current state-of-the-art, DeepDSC. SiamCDR$_{RF}$ also identifies FDA-approved therapies and recommends drug repurposing candidates with reasonable success. Notably, via pathway analysis of genes significantly correlated with SiamCDR$_{RF}$'s drug prioritization, we find significant enrichment of pathways known to induce resistance. This implies the proposed model has learned to tailor its predictions based on transcriptomic signals of drug resistance. Finally, we present 19 drug repurposing candidates for the treatment of either BC, HNC, GC, or PC. Eight of these drugs are unique to an individual cancer, further demonstrating SiamCDR$_{RF}$'s tailored prioritization.

Despite its extremely promising performance, SiamCDR$_{RF}$, like other deep learning models, would likely benefit from additional data. Drug representations may be further enhanced by leveraging inhibition/activation information while pretraining $Enc_d$. As observed in **Figure 4d**, by using only gene target to determine drug similarity, agonists and antagonists are grouped together in the embedding space. Cell line representations may also be improved with the use of additional omics types. This is because each omics type possesses distinct information that may further enhance the model's predictive power. Finally, producing pan-cancer drug recommendations is difficult problem as drug sensitivities vary both within and across cancer types. Given that drug sensitivities among cells of the same cancer are generally more similar than drug sensitivities of cells of different cancer types, a mixture of experts may improve SiamCDR$_{RF}$'s recommendations by enabling each 'expert' model to focus only on predicting drug sensitivities for cells grouped by cancer type.

**Methods**:

Drug-cell line pairs: Cancer drug response (CDR) data was obtained from the Broad Institute (PRISM Repurposing 19Q4's secondary screen)[30]. In total, this data set contains information on 701,004 drug-cell line pairs. Data was collected from chemical-perturbation viability screens for serial dilutions of 1,448 small-molecule drugs against 480 cell lines. Dose-response curves were fit to the viability screens and used to predict drug response-related metrics. The CDR was calculated from features of the dose-response curve via a custom effective score as will be discussed later in Equation (1). As the score was calculated from AUC, lower limit, and $IC_{50}$ values, pairs missing this data were excluded. As lower limit is the minimum cell viability a drug can achieve, pairs with predicted lower limits $< 0$ were excluded. Additionally, we excluded drug-cell line pairs with $R^2 < 0.7$ to ensure the dose-response curves were well fit to the data. We further processed data by removing duplicate pairs with the following process: 1) in accordance with PRISM documentation, we retained cell lines with the MTS010 screen ID as these are higher quality screens; and 2) if none of the pair's duplicates were from the MTS010 screen, the duplicate with the highest $R^2$ was retained. We then removed 1) previously withdrawn drugs and 2) cell lines matching any of the following criteria: a) less than 1% of the cell line's screened drugs were highly effective (*CES* $\geq 7.2734$; see "Effective score"); b) the cell line's cancer type was unknown; and c) the cell line's RNA-seq was unavailable (see "Cell line gene expression"). These criteria ensure that, for each cell line, a minimum number of effective drugs were observed to learn to predict CDR and that any drug identified as high priority candidates would be actionable. After removing pairs with missing or low-quality data, we were left with 67,838 drug-cell line pairs from 1,105 drugs and 419 cell lines. Drugs and cell lines are initially represented as 256-bit Morgan fingerprints[4,44] and RNA expression of a subset of cancer genes (see "Cell line gene expression") and are denoted by vectors **f** and **g,** respectively.

Effective score: $IC_{50}$, or the half-maximal inhibitory concentration, represents the concentration of a drug needed to reach 50% cell viability and partially conveys drug effectiveness. However, this measure only evaluates drug potency, completely ignoring both drug efficacy and the minimum viability a drug can achieve. Two drugs with similar $IC_{50}$ would be considered similar even if one reached 5% viability and the other, 49%. To better balance potency and efficacy, we propose a custom effective score (CES):

$$\text{Equation (1):} \quad CES \ = \ \log \left( \frac{\text{AUC} + \text{lower limit} + IC_{50}}{2 \times \text{AUC} \times \text{lower limit} \times IC_{50}} \right).$$

AUC for dose-dependent curves conveys a drug's cumulative effect across treatment concentrations by incorporating both potency and effectiveness. Specifically, lower AUC values indicate higher cellular sensitivity to treatment, even at low concentrations. Lower limit is the minimum possible viability a drug can elicit. Including this term will give greater importance to drugs that are more likely to completely eradicate all cells of a cancer. The score is binarized—threshold $\approx 7.27(\mu + 1.28\sigma)$—such that those with effective scores in the top-10% are labeled 1 and 0 otherwise. Using these labels allow models to be trained to identify only the most effective drugs for each cell line.

Cell line gene expression: We produce cell line representations by leveraging gene expression data obtained from the Broad Institute (DepMap Public 22Q2)[45]. Gene expression was leveraged as it is both information-rich and the most widely available of any omics data type, making our model and its predictions more accessible. The expression of 18,964 protein coding genes was measured across 1,406 cell lines (473 of which were evaluated for their CDR; see "Drug-cell line pairs"). Using KEGG's documented cancer pathways[46–48], we considered the expression for only 463 cancer-related genes (**Supplementary Note 1**). We excluded any remaining cell lines, not evaluated for CDR data, if 1) their cancer types had fewer than 10 cell lines with expression data or 2) their listed cancer type was either 'Unknown' or 'Non-cancerous'. This yielded 864 cell lines, which we used to pretrain DeepDSC's autoencoder[3] and SiamCDR's cell line Siamese neural network (SNN) (see "Learning cell line and drug representations"). The distribution of cancers among the pretraining data is presented in **Supplementary Table 1a**.

The 419 cells lines with CDR data were used for model training, validation, and testing. Cancers evaluated on fewer than 15 cell lines had all their cell lines (67) reserved as a test set; we denote this subset of the data novel cancers, to evaluate model generalizability on novel cancers. **Supplementary Table 1b** reports the distribution of cell lines by-cancer among the novel cancer test set. The remaining cell lines underwent random training/test splitting. To preserve the distribution of cancer across each split, random sampling was conducted by cancer. Specifically, 15% of the cell lines from each cancer were reserved for testing model generalizability on novel cell lines with cancers seen during training; we denote this subset of the data trained-on cancers. The remaining cell lines of each cancer were randomly assigned to one of five folds in approximately equal proportions to be used for 5-fold cross-validation. The distribution of cell lines by cancer in the training folds and trained-on test set is reported in **Supplementary Table 1c**.

Baseline method: We evaluate our proposed framework by comparing the performance of models trained from it against a state-of-the-art method: DeepDSC[3]. We select this method for its good performance relative to other published methods and its use of both gene expression and drugs as input. As such, improvements in performance garnered by models produced by our proposed framework will not be due to different data used. We do not compare against RefDNN, despite its use of transcriptomics, as it requires a reference drug set for which the response is known for all cell lines. There are no drugs for which this is the case in our CDR data set, nor in real applications is this likely to be the case. Other models such as DeepCDR[49], and DeepDR[50] may achieve comparable performance to DeepDSC; however, these also utilize multi-omics data sets. Doing so provides the models with additional information about cell lines that is unavailable when using a single omics data type. This precludes fair comparison of methods as it becomes impossible to ascertain whether differences in performance arise from methodological differences or the use of distinct information.

Overview of SiamCDR framework: **Figure 1a** highlights the distinct components comprising the proposed SiamCDR framework. Solid boxes are consistent across each variation of the framework, while boxes with dashed borders represent aspects evaluated during hyperparameter tuning. Drug-cell line pairs are input to the framework. Each drug is initially represented by a vector **f**; each cell line is represented by a vector **g**. Depending on the model, one or both of drugs and cell lines are projected to an embedding space using a DNN encoder (*Enc*) pretrained by a Siamese neural network (SNN) (see "*L*earning cell line and drug representations" and **Figure 1b**). We denote $Enc_d$ and $Enc_c$ as the drug and cell line encoders, respectively, with $Enc_d$ and $Enc_c$ projecting **f** to $\mathbf{e}_d$ and **g** to $\mathbf{e}_c$, respectively. Here, $\mathbf{e}_d$ and $\mathbf{e}_c$ are vectors that represent drug and cell line embeddings, respectively. Embedding drugs and cell lines in this way enhances the framework's ability to capture the most salient information for CDR predictions, imparting higher expressiveness to the embeddings produced. Drug and cell line representations are then combined and input to an end classifier (see "SiamCDR's end classifier") which is used to predict the relative CDR for each drug-cell line pair.

Learning cell line and drug representations: Our proposed SiamCDR framework leverages a SNN[5] to pretrain $Enc_d$ and $Enc_c$ to encode drugs or cell lines via contrastive learning, respectively. The SNN structure is illustrated in **Figure 1b**. During training, an *Enc* will be applied to a pair of inputs, producing an embedding for each. Note the *Enc*'s weights are shared when embedding each input. The SNN then applies a sigmoid activation function to the Euclidean distance between the embeddings, producing a probability that the inputs are from different groups. Ground-truth labels indicate if the inputs are from different groups (1) of the same group (0). Binary cross entropy loss is optimized such that *Enc* preserves intra- and inter-group similarity relationships within the embedding space. By doing this, members of the same group are mapped close together and those of different groups are mapped further apart, allowing us to explicitly capture task-specific information.

For our context, we employ nearly identical training structures when pretraining $Enc_d$ and $Enc_c$ to embed drugs and cell lines. The only differences are the input and how the ground-truth labels are determined. Drugs are grouped by their gene targets; cell lines are grouped by cancer types. $Enc_d$, which produce $\mathbf{e}_d$, is pretrained on the subset of drugs in the drug-cell line pair data with at least one reported gene target. $Enc_c$, which produce $\mathbf{e}_c$, is pretrained on 864 cells not evaluated for CDR. Hyperparameters were tuned via an exhaustive search; the best hyperparameters for both $Enc_d$ and $Enc_c$ are indicated in **Supplementary Table 2.**

SiamCDR's end classifiers: We evaluate the following classifiers: DNN, logistic regression (LR), random forest (RF). DNNs are extremely popular because of their flexibility and capacity to learn complex, non-linear patterns. However, this capacity comes as the cost of many learnable parameters, increasing the risk of overfitting when training data is limited. On the other hand, LR models are simple models that make predictions based on linear combinations of features. Thus, while they benefit from having fewer trainable parameters than DNN, making them less prone to overfitting, they fail to capture non-linear patterns. Finally, RF models are ensemble methods comprised of multiple decision trees with each tree trained on a bootstrap-sampled subset of the training data. Its ensemble nature, which relies on consensus among trees, provides greater stability to predictions. Furthermore, the risk of overfitting can be mitigated by controlling tree depth and the minimum number of observations in each split. The complexity of the entire RF model is limited—compared to DNNs—by the simplicity each tree itself. However, by making a series of nested, linear decision boundaries, these trees can capture more complex relationships than LR models. Hyperparameter tuning options for RF and DNN classifiers are reported in **Supplementary Tables 3a** and **b**, respectively.

Model selection: We evaluate the performance of all combinations of the overall framework discussed above and present the results in **Supplementary Tables 4** and **5**. From these results, we select the top-performing architecture for each end classifier with respect to $P_{cell}@k$ and $P_{cancer}@k$ for trained-on cancers. The best model for the DNN, RF, and LR end classifiers is denoted SiamCDR$_{DNN}$, SiamCDR$_{RF}$, and SiamCDR$_{LR}$, respectively. All three models leverage learned cell line representation ($\mathbf{e}_c$). SiamCDR$_{RF}$ represents drugs with learned embeddings ($\mathbf{e}_d$), while SiamCDR$_{DNN}$ and SiamCDR$_{LR}$ use Morgan fingerprints ($\mathbf{f}$).

Evaluation metrics: We evaluate model performance via precision@$k$ for both cell lines and cancers (Equations (2) and (3), respectively), which measures the proportion of highly effective drugs among a model's top-$k$ prioritized drugs. Each metric provides a different level of granularity with which to evaluate the precision of model prioritizations. For each model, we obtain a cell line's prioritization ordering for each cell line by sorting its predicted CDR scores for candidate drugs in descending order. The nearer the top a drug is, the higher a model has prioritized the drug a given cell line.

Equation (2): $P_{cell}@k(i) \ = \ \frac{D_i(k) \cap E_i}{D_i(k)}.$

Equation (3):  $P_{\text{cancer}}@k(j) = \frac{\sum_{i=1}^{N_j} P_{cell}@k(i)}{N_j}$ .

In Equation (2), $D_i(k)$ is the set of top-$k$ prioritized drugs for the $i$-th cell line and $E_i$ is the set of all effective drugs for that cell line. In Equation (3), $N_j$ is the set of cell lines of the $j$-th cancer. In both cases, higher $P_{\text{cell}}@k$ $and$ $P_{\text{cancer}}@k$ scores indicate a model is better able to prioritize effective drugs at the very top.

<u>Statistical analysis</u>: Significance of the performance of top-3 best SiamCDR models (SiamCDR<sub>LR</sub>, SiamCDR<sub>RF</sub>, and SiamCDR<sub>DNN</sub>) compared to the DeepDSC baseline across 5-fold cross-validation was obtained via Bonferroni multiple-hypothesis corrected, pairwise, two-tailed, independent t-tests.

<u>Feature importance</u>: We measure feature importance to discern their individual contributions to model predictions. Higher importance reflects greater contribution. In our context, we anticipate high-performance models to leverage both drug and cell line information when making predictions, ranking features derived from both drugs and cell lines highly with respect to their importance. Random forest models (e.g., SiamCDR<sub>RF</sub>) provide feature importance as a native attribute. For DNNs (e.g., SiamCDR<sub>DNN</sub> or DeepDSC), we estimate individual feature importance as the mean magnitude of SHAP[51] values calculated across 5,000 randomly sampled training examples. Finally, we estimate feature importance for logistic regression classifiers (e.g., SiamCDR<sub>LR</sub>) by the magnitude of their coefficients. However, we first evaluate coefficient stability as unstable coefficients serve as poor indicators of feature importance. Coefficient stability reflects consistency in model feature preferences and can be assessed by measuring the variance in coefficients obtained across multiple training folds. High stability is indicated by a small variance relative to the average magnitude of coefficients and *vice versa*. We find the mean variance of SiamCDR$_{LR}$'s coefficients to be 2-orders of magnitude smaller than the mean value of SiamCDR$_{LR}$'s coefficients ($7.99 \times 10^{-3}$ and 0.38, respectively), suggesting that SiamCDR$_{LR}$'s coefficients can be reliably used to estimate feature importance.

Clustering: We use the t-SNE method[52] to project the drug and cell line embedding spaces to two-dimensional spaces. This facilitates a qualitative evaluation of the expressiveness of embeddings used by DeepDSC and SiamCDR$_{RF}$. Proximity within the projected spaces is positively associated with embedding similarity. As such, we expect embeddings with high expressiveness to produce well-defined clusters comprised of drugs or cell lines of the same (or similar) MOA or cancer, respectively.

Embedding expressiveness: We assess the expressiveness of our learned drug and cell line representations using intra-group similarity and inter-group separability. Groups are defined as MOAs for drugs and cancer types for cell lines. In other words, these metrics evaluate how well drug representations cluster by MOA and how well cell line representations cluster by cancer type. We define intra-group similarity as the average pairwise cosine similarity between all members of a group and inter-group similarity as the average pairwise cosine similarities of each group member to all non-group members. Both range from -1 to 1 with more positive values denoting increased similarity and more negative values associated with increased dissimilarity. We gauge the distinctiveness of a group's cluster from those of other groups via the ratio between its intra- and inter-group similarity, which we denote 'inter-group separability'. Higher inter-group separability indicates higher degrees of separation between groups. In our context, embeddings with high expressiveness should be capable of discerning drugs of different MOAs and cell lines of different cancer types. As such, they will, ideally, attain high values with respect to both metrics.

**Code and data availability:**

All the processed data, code, and instructions to replicate our work is freely available here: https://github.com/ninglab/SiamCDR. Additionally, the source and versions of our work's primary software, hardware, and data dependencies is presented in **Supplementary Table 11**.

**Acknowledgements:** This project was made possible, in part, by support from the National Science Foundation grant no. IIS-2133650 (X.N.). Any opinions, findings, and conclusions or recommendations expressed in this paper are those of the authors and do not necessarily reflect the views of the funding agency.

**Author contributions:** Conceptualization, P.L and X.N.; methodology, P.L. and X.N.; software, P.L.; validation, P.L. and X.N.; formal analysis, P.L. and X.N.; investigation, P.L. and X.N.; resources, X.N.; data curation, P.L.; writing – original draft, P.L.; writing – review & editing, P.L. and X.N.; visualization, P.L.; supervision, X.N.; project

administration, X.N.; funding acquisition, X.N.

**Competing interests:**

The authors declare no competing interests.

**Figures:**

**Figure 1 Model architectures.**

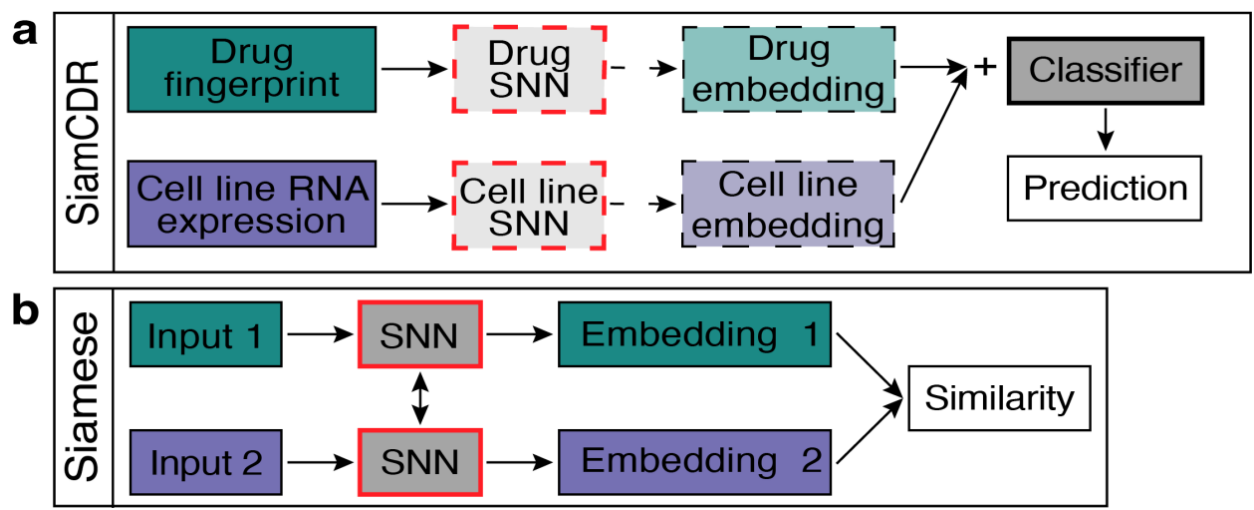

Depicts architectures for components proposed by this work: **a**) siamese neural network and

**b**) SiamCDR. For both**,** boxes with bold borders and a grey face denote trained components. The input pair in **a** is either a pair of drugs or pair or cell lines depending on if drug or cell line encoder is being trained. Dashed lines and box borders in **b** indicate optional components by variation. See respective sections in Methods for complete details.

**Figure 2 Comparing effective score to model predictions.**

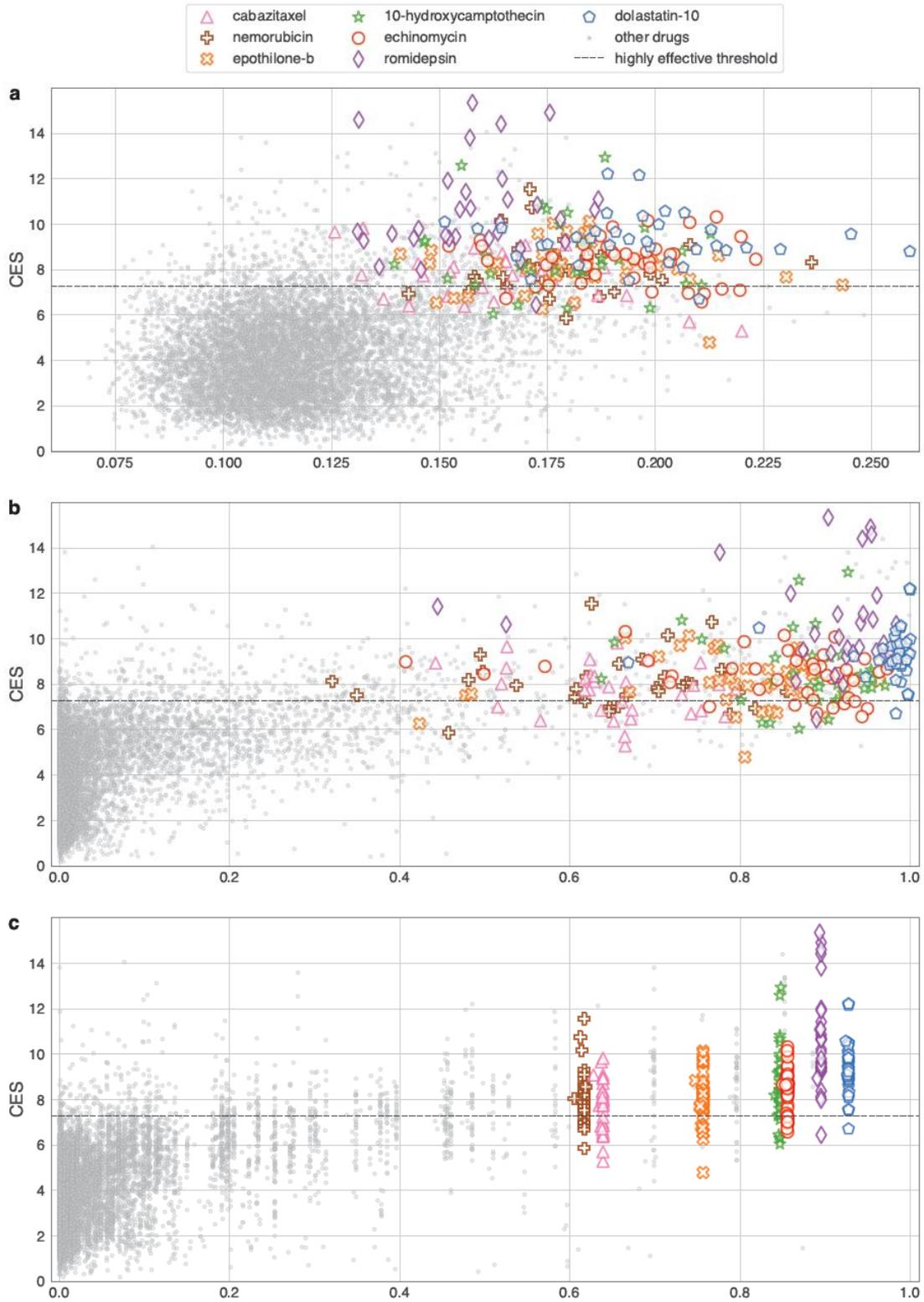

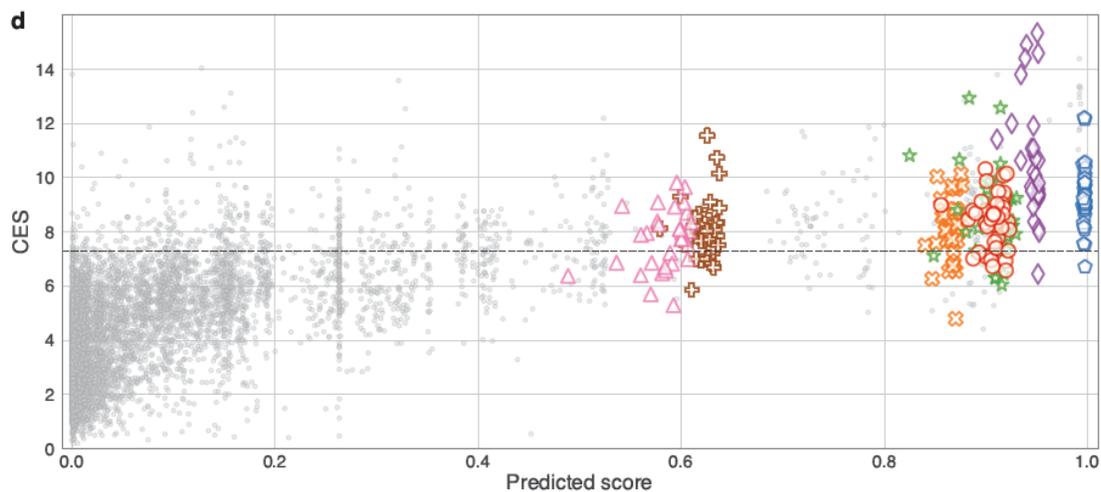

Plots the relationship between CES (continuous effective score) and scores predicted by

**a**) DeepDSC, **b**) SiamCDR$_{RF}$, **c**) SiamCDR$_{LR}$, and **d**) SiamCDR$_{DNN}$. Each point represents a drug-cell line pair. Drug-cell line pairs containing drugs recommended in the top-3 for at least 3 cell lines by all 4 models are highlighted with distinct colors and shapes (see legend). Note the scale of predicted score in **a** is different than **b-c**. This was done to allow the general trend to be visualized.

**Figure 3 Scaled drug- and cell line-derived feature importance for model predictions.**

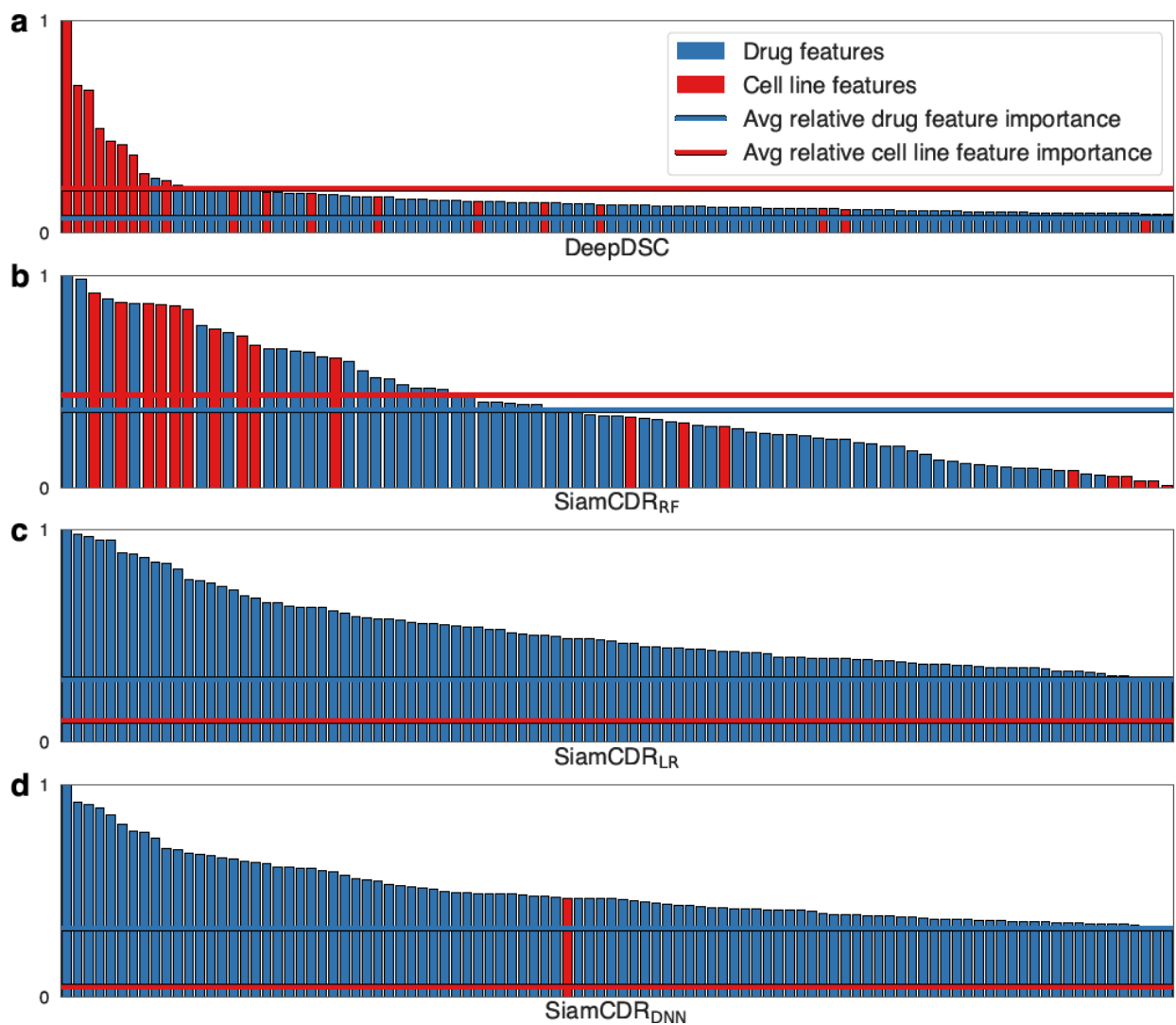

Minmax scaled feature importance (> 0.01) for the top-100, non-zero features is plotted in descending order along the x-axis for **a**) DeepDSC, **b**) SiamCDR$_{RF}$, **c**) SiamCDR$_{LR}$, and **d**) SiamCDR$_{DNN}$. In each plot, the average relative feature importance for both drug- and cell line-derived features is plotted with horizonal lines.

**Figure 4 t-SNE plots for cell line and drug feature representations.**

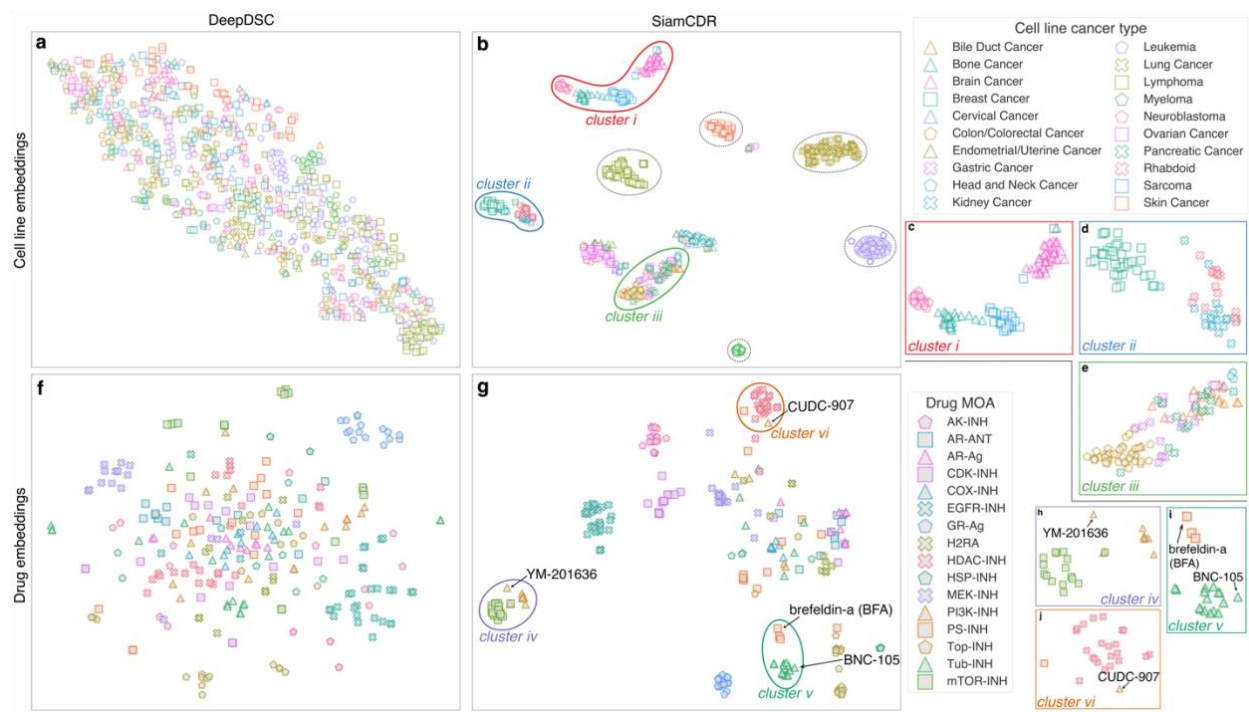

T-SNE plots were from the embeddings produced by both **a**) DeepDSC's autoencoder and **b**) SiamCDR's cell line encoder for cancers with at least 15 cells. Each cancer is represented by a distinct color/shape combination. In **b**, clusters of single cancer types are highlighted with dotted, black lines; clusters discussed in the Results are highlighted and labeled with distinct colored lines. *Clusters i, ii, and iii* in **b** are magnified in **c**, **d**, and **e**, respectively. We also produce t-SNE plots for drugs with MOAs with at least 10 drugs in the pre-training data using either **f**) 256-bit Morgan fingerprints (DeepDSC) or **g**) SiamCDR$_{RF}$'s drug encoder embeddings. Each MOA is represented by unique color/shape combination. Clusters discussed in the Results section are highlighted in **g** and labeled with distinct colored lines. Clusters *iv, v,* and *vi* in **g** are magnified in **h**, **i**, and **j**, respectively. The axis scales for all plots have been adjusted to best fit the data. Abbreviations: PS: protein synthesis, TOP: topoisomerase, H2RA: histamine receptor antagonist, TP: tubulin polymerization, AR: adrenergic receptor, COX: cyclooxygenase, GR: glucocorticoid receptor, AK: aurora kinase, -INH: inhibitor, -A: antagonist, -Ag: agonist.

**Tables:**

**Table 1a** $P_{\text{cell}}@k$ **for cell lines with trained-on cancers.**

| Model | $P_{\text{cell}}@1$ | $P_{\text{cell}}@2$ | $P_{\text{cell}}@3$ | $P_{\text{cell}}@4$ | $P_{\text{cell}}@5$ | $P_{\text{cell}}@10$ |
|---|---|---|---|---|---|---|
| DeepDSC | 0.7059 | 0.6000 | 0.5608 | 0.5392 | 0.5294 | 0.5031 |
| SiamCDR$_{\text{LR}}$ | 0.9519 | *0.9368 | *0.9194 | **0.8829 | **0.8641 | ***0.8007 |
| SiamCDR$_{\text{RF}}$ | **0.9647** | *0.9196 | *0.8915 | **0.8657 | **0.8486 | ***0.8185 |
| SiamCDR$_{\text{DNN}}$ | 0.9490 | *0.9137 | *0.8941 | *0.8598 | **0.8353 | ***0.8405 |

Significance levels ($\alpha \leq 0.1, 0.05, 0.01$; indicated by *, **, ***, respectively) are determined from *p*-values obtained via Bonferroni correction (n=4) of two-tailed t-tests comparing the performances of SiamCDR$_{\text{LR}}$, SiamCDR$_{\text{LR}}$, and SiamCDR$_{\text{LR}}$ against DeepDSC.

**Table 1b** $P_{\text{cell}}@k$ **for cell lines with novel cancers.**

| Model | $P_{\text{cell}}@1$ | $P_{\text{cell}}@2$ | $P_{\text{cell}}@3$ | $P_{\text{cell}}@4$ | $P_{\text{cell}}@5$ | $P_{\text{cell}}@10$ |
|---|---|---|---|---|---|---|
| DeepDSC | 0.7938 | 0.6846 | 0.6164 | 0.6023 | 0.5742 | 0.5463 |
| SiamCDR$_{\text{LR}}$ | *0.9508 | 0.9123 | *0.9046 | *0.8738 | **0.8622 | ***0.8059 |
| SiamCDR$_{\text{RF}}$ | **0.9569 | *0.9123 | 0.8892 | *0.8554 | *0.8363 | ***0.8090 |
| SiamCDR$_{\text{DNN}}$ | **0.9723** | *0.9354** | **0.9036** | *0.8792** | **0.8591** | ***0.8239** |

Significance levels ($\alpha \leq 0.1, 0.05, 0.01$; indicated by *, **, ***, respectively) are determined from *p*-values obtained via Bonferroni correction (n=4) of two-tailed t-tests comparing the performances of SiamCDR$_{\text{LR}}$, SiamCDR$_{\text{LR}}$, and SiamCDR$_{\text{LR}}$ against DeepDSC.

**Table 2a** $P_{\text{cancer}}@k$ **for cell lines with trained-on cancers.**

| Model | $P_{\text{cancer}}@1$ | $P_{\text{cancer}}@2$ | $P_{\text{cancer}}@3$ | $P_{\text{cancer}}@4$ | $P_{\text{cancer}}@5$ |
|---|---|---|---|---|---|
| DeepDSC | 0.7182 | 0.6083 | 0.5640 | 0.5397 | 0.5347 |
| SiamCDR$_{\text{LR}}$ | 0.9571 | ***0.9435** | ***0.9179** | ***0.8820** | ***0.8597** |
| SiamCDR$_{\text{RF}}$ | ***0.9712** | *0.9313 | *0.8968 | **0.8695 | **0.8485 |
| SiamCDR$_{\text{DNN}}$ | 0.9503 | *0.9225 | *0.9012 | *0.8674 | **0.8436 |

Significance levels ($\alpha \leq 0.1, 0.05, 0.01$; indicated by *, **, ***, respectively) are determined from *p*-values obtained via Bonferroni correction (n=4) of two-tailed t-tests comparing the performances of SiamCDR$_{\text{LR}}$, SiamCDR$_{\text{LR}}$, and SiamCDR$_{\text{LR}}$ against DeepDSC. Precision@k for each cancer is presented in **Supplementary Table 6**.

**Table 2b** $P_{\text{cancer}}@k$ **for cell lines with novel cancers.**

| Model | $P_{\text{cancer}}@1$ | $P_{\text{cancer}}@2$ | $P_{\text{cancer}}@3$ | $P_{\text{cancer}}@4$ | $P_{\text{cancer}}@5$ |
|---|---|---|---|---|---|
| DeepDSC | 0.8157 | 0.7155 | 0.6447 | 0.6174 | 0.5793 |
| SiamCDR$_{\text{LR}}$ | *0.9621 | 0.9283 | ***0.9267** | ***0.8821** | **0.8504 |
| SiamCDR$_{\text{RF}}$ | *0.9616 | *0.9311 | 0.8840 | 0.8585 | ***0.8367** |
| SiamCDR$_{\text{DNN}}$ | ***0.9794** | ***0.9547** | 0.9237 | *0.8817 | **0.8591 |

Significance levels ($\alpha \leq 0.1, 0.05, 0.01$; indicated by *, **, ***, respectively) are determined from *p*-values obtained via Bonferroni correction (n=4) of two-tailed t-tests comparing the performances of SiamCDR$_{\text{LR}}$, SiamCDR$_{\text{LR}}$, and SiamCDR$_{\text{LR}}$ against DeepDSC. Precision@k for each cancer is presented in **Supplementary Table 6**.

## Table 3 DeepDSC and SiamCDR$_{RF}$'s by-cancer prioritizations of FDA-approved drugs.

| Cancer | Cell line count | Drug count | Mean priority | | | Max priority | | | Highest prioritized drug (%) | |
|---|---|---|---|---|---|---|---|---|---|---|
| | | | DSC | CDR | α | DSC | CDR | α | DSC | CDR |
| Bladder | 17 | 3 | 22.19 | 98.92 | **** | 5.65 | 20.71 | **** | doxorubicin (100.0) | valrubicin (64.7) |
| Breast | 43 | 17 | 133.14 | 163.68 | **** | 2.00 | 2.72 | **** | docetaxel (62.8) | gemcitabine (100.0) |
| Cervical | 18 | 1 | 10.00 | 104.94 | **** | 10.00 | 104.94 | **** | topotecan (100.0) | topotecan (100.0) |
| Colorectal | 50 | 9 | 183.50 | 238.18 | **** | 10.98 | 16.1 | **** | irinotecan (100.0) | SN-38 (100.0) |
| E/U | 22 | 1 | 222.27 | 332.27 | **** | 227.27 | 332.27 | **** | MTX (100.0) | MTX (100.0) |
| Gastric | 39 | 3 | 144.81 | 159.10 | ** | 2.13 | 25.54 | **** | docetaxel (100.0) | docetaxel (100.0) |
| **HNC** | 37 | 2 | 168.86 | 33.27 | **** | 2.11 | 25.86 | **** | docetaxel (100.0) | docetaxel (86.5) |
| Kidney | 33 | 7 | 84.71 | 204.76 | **** | 5.64 | 8.70 | **** | doxorubicin (100.0) | temsirolimus (84.8) |
| **Leukemia** | 104 | 33 | 190.98 | 169.65 | **** | 5.62 | 4.47 | **** | doxorubicin (100.0) | vincristine (89.4) |
| Liver | 9 | 1 | 118.33 | 290.11 | **** | 118.33 | 290.11 | **** | sorafenib (100.0) | sorafenub (100.0) |
| Lung | 118 | 19 | 134.55 | 169.89 | **** | 2.00 | 2.58 | **** | docetaxel (91.5) | gemcitabine (99.2) |
| **Lymphoma** | 83 | 12 | 176.66 | 103.43 | **** | 5.14 | 3.82 | **** | doxorubicin (65.1) | romidepsin (100.0) |
| **Myeloma** | 30 | 9 | 143.19 | 73.22 | **** | 5.67 | 6.20 | * | doxorubicin (100.0) | ixazomib (90.0) |
| **Neuroblastoma** | 29 | 1 | 271.28 | 67.10 | **** | 271.28 | 67.10 | **** | cytoxan (100.0) | cytoxan (100.0) |
| **Ovarian** | 38 | 7 | 176.88 | 78.73 | **** | 2.79 | 2.71 | | paclitaxel (100.0) | gemcitabine (100.0) |
| Pancreatic | 24 | 7 | 193.88 | 226.23 | **** | 21.96 | 2.54 | **** | everolimus (100.0) | gemcitabine (100.0) |
| **Prostate** | 10 | 5 | 77.78 | 35.66 | **** | 2.40 | 2.80 | | docetaxel (100.0) | cabazitaxel (100.0) |
| Sarcoma | 35 | 2 | 101.10 | 140.20 | **** | 58.97 | 30.86.93 | **** | vinblastine (100.0) | vinblastine (100.0) |
| **Skin** | 51 | 6 | 235.68 | 174.33 | **** | 112.96 | 39.82 | **** | vindesine (100.0) | dabrafenib (100.0) |
| Thyroid | 7 | 5 | 205.01 | 286.19 | **** | 89.53 | 204.73 | **** | lenvatinib (100.0) | cinacalcet (60.0) |

The average mean and max priority of indicated drugs across cell lines with a given cancer (**Supplementary Table 9**) are presented. Bolded cancers indicate those for which SiamCDR$_{RF}$ achieves higher mean priorities than DeepDSC. Significance levels ($\alpha \leq 0.1, 0.05, 0.01, 0.001$; indicated by *, **, ***, **** respectively) are determined from $p$-values obtained from a two-tailed, independent t-test and denotes the significance of SiamCDR$_{RF}$'s prioritizations (CDR) compared to DeepDSC's prioritizations (DSC). (%) in highest prioritized drug indicates the % of cell lines prioritizing the listed drug highest among their FDA-approved indications. Cancer and drug abbreviations – CC: Colon/Colorectal; E/U: Endometrial/Uterine; HNC: Head and Neck; MTX: methotrexate; Cytoxan: cyclophosphamide.

**Table 4 Highly prioritized drugs as candidates for repurposing.**

| Cancer type: FDA-approved targeted therapies | Drug | Avg Rank | MOA | Gene targets | Indication | Evidence |
|---|---|---|---|---|---|---|
| | valrubicin | 70.6 | DNA-INH TOP-INH | TOP2A | A (BC) | ----- |
| | triptolide | 2.1 | RNA-POL-INH | CYP2C19 RELA | P3 (KD) | 25,26 |
| | HCPT | 4.8 | TOP-INH | TOP1 | preclinical | 25 |
| | dol-10 | 5.0 | TP-INH | TUBB | P2 (L / Ly) | |
| | romidepsin | 11.5 | HDAC-INH | CYP2B6 CYP2C19 CYP3A5 HDAC[1-9] | A (Ly) | 53,54 |
| | YM-155 | 12.1 | SUR-INH | BIRC5 | P2 (PC, M, Ly) | 55 |
| Bladder atezolizumab, avelumab, enfortumab, erdafitinib, nivolumab, pembrolizumab, sacituzumab govitecan-hziy | cabazitaxel | 13.4 | miTUB-INH | CYP2C8 CYP3A5 SLCO1B3 TUBA1[A-B] TUBA3[D-E] TUBA4A TUBB, TUBB[1,6,8] TUBB2[A-B] TUBB4A | A (PC) | 22 |
| | gemcitabine | 15.3 | RR-INH | CMPK1 RRM[1-2] TYMS | A (BrC, NSCLC, OC, PaC) | 26,56 |
| | GZD824 | 20.0 | BAK-INH | ABL1 BCR | preclinical | 24 |
| | OTS167 | 25.6 | MELZK-INH | MELK | P1 (L) | 57 |
| | Oltipraz | 26.1 | NRF2 activator | ANG CYP2B6 NFE2L2 | P3 (NAFLD) | 23 |
| | CUDC-907 | 30.0 | PI3K-INH | HDAC2 PIK3R1 | P3 (Ly) | 58 |
| | temsirolimus | 37.2 | mTOR-INH | MTOR PTEN | A (RCC) | 59 |
| | ixazomib | 42.0 | Ptsm-INH | | A (My) | 60 |
| | docetaxel | 85.1 | TP-INH | BLC2 MAP[2, 4, T] NR1I2 TUBA1[A-C] TUBA3[C-E] TUBA4A TUBB TUBB2[A-B] TUBB[3, 6, 8] TUBB4[A-B] | A (BrC, NSCLC, PC, GC, HNC) | ----- |
| HNC cetuximab, nivolumab, pembrolizumab | YM-155 | 8.9 | SUR-INH | BIRC5 | P2 (PC, M, Ly) | 33 |
| | BGT226 | 23.6 | PI3K-INH | MTOR PIK3CA PIK3CB PIK3CG | P1/2 (BrC) | 34 |
| | OTS167 | 26.3 | MELZK-INH | MELK | P1 (L) | 61 |

| | | | | | |
|---|---|---|---|---|---|
| | temsirolimus | 33.4 | mTOR-INH | MTOR<br>PTEN | A (RCC) | 32 |
| | ninlaro | 37.3 | Ptsm-INH | | A (My) | 31 |
| | litronesib | 37.4 | KLSP-INH | KIF11 | P2 (BrC) | 62 |
| | SN-38 | 42.5 | TOP-INH | TOP1 | A (CC) | 35 |
| | poziotinib | 45.9 | EGFR-INH | EGFR<br>ERBB[2, 4] | P2 (LC) | 29 |
| | <u>docetaxel</u> | 79.4 | TP-INH | BLC2<br>MAP[2, 4, T]<br>NR1I2<br>TUBA1[A-C]<br>TUBA3[C-E]<br>TUBA4A<br>TUBB<br>TUBB2[A-B]<br>TUBB[3, 6, 8]<br>TUBB4[A-B] | A (BrC, NSCLC, PC, GC, HNC) | ----- |
| | triptolide | 2.3 | RNA-POL-INH | CYP2C19<br>RELA | P3 (KD) | 27 |
| | ExM | 7.1 | TOP-INH | TOP1 | P3 (PaC) | 37 |
| <u>Gastric</u><br><br>fam-trastuzumab,<br>nivolumab,<br>pembrolizumab,<br>ramucirumab,<br>trastuzumab | romidepsin | 9.3 | HDAC-INH | CYP2B6<br>CYP2C19<br>CYP3A5<br>HDAC[1-9] | A (Ly) | 39 |
| | YM-155 | 11.7 | SUR-INH | BIRC5 | P2 (PC, M, Ly) | 40 |
| | alvespimycin | 13.1 | HSP-INH | HSP90AA1 | P2 (BrC) | 63 |
| | cabazitaxel | 15.6 | miTUB-INH | CYP2C8<br>CYP3A5<br>SLCO1B3<br>TUBA1[A-B]<br>TUBA3[D-E]<br>TUBA4A<br>TUBB,<br>TUBB[1,6,8]<br>TUBB2[A-B]<br>TUBB4A | A (PC) | 64 |
| | BGT226 | 21.3 | PI3K-INH | MTOR<br>PIK3CA<br>PIK3CB<br>PIK3CG | P1/2 (BrC) | |
| | GZD824 | 22.0 | BAK-INH | ABL1<br>BCR | preclinical | 65 |
| <u>Prostate</u><br><br>abiraterone acetate,<br>apalutamide,<br>cabazitaxel,<br>darolutamide,<br>enzalutamide,<br>lutetium Lu 177<br>vipivotide tetraxetan, | <u>cabazitaxel</u> | 17.7 | miTUB-INH | CYP2C8<br>CYP3A5<br>SLCO1B3<br>TUBA1[A-B]<br>TUBA3[D-E]<br>TUBA4A<br>TUBB,<br>TUBB[1,6,8]<br>TUBB2[A-B]<br>TUBB4A | A (PC) | ----- |
| | dol-10 | 2.1 | TP-INH | TUBB | P2 (L / Ly) | 43 |
| | triptolide | 3.3 | RNA-POL-INH | CYP2C19<br>RELA | P3 (KD) | 28 |
| | HCPT | 6.3 | TOP-INH | TOP1 | precinical | 42 |
| | ExM | 8.0 | TOP-INH | TOP1 | P3 (PaC) | 66 |

| olaparib, talazoparib tosylate, radium 223 dichloride, rucaparib camsylate | alvespimycin | 8.8 | HSP-INH | HSP90AA1 | P2 (BrC) | 38 |
| | YM-155 | 9.1 | SUR-INH | BIRC5 | P2 (PC, M, Ly) | |
| | camptothecin | 13.6 | TOP-INH | TOP1 | P3 (CC) | 41 |
| | gemcitabine | 16.1 | RR-INH | CMPK1 RRM[1-2] TYMS | A (BrC, NSCLC, OC, PaC) | 67 |

Abbreviations by column—**Drug**) HCPT: 10-hydroxycamptothecin, dol-10: dolastatin-10, ExM: exatecan-mesylate, ninlaro: ixazomib citrate; **MOA**) -INH: inhibitor, -A: antagonist, -Ag: agonist, -SA: stabilizing agent, TOP: topoisomerase, RNA-POL: RNA polymerase, TP: tubulin polymerization, HSP: heat shock protein, MEK: mitogen-activated protein kinase, BCWS: bacterial cell wall synthesis, PI3K: phosphatidylinositol-3 kinase, miTUB: microtubule, RR: ribonucleotide reductase, BAK: Bcr-Abl kinase, ER: estrogen receptor, GR: glucocorticoid receptor, PR: progesterone receptor, KLSP: kinesin-like spindle protein, MELZK: maternal embryonic leucine zipper kinase, Ptsm: proteasome; **Gene targets**) Multiple genes that are identical except for last character are collapsed using square brackets. Within the brackets, the final letters or numbers are either comma delimited or indicated with ranges. For example, HOX[1-3, 6] would indicate that the drug targets HOX1, HOX2, HOX3, and HOX6; **Indication**) BC: bladder cancer, BrC: breast cancer, CC: colorectal cancer, HNC: head and neck cancer, KD: kidney disease, L: leukemia, LC: lung cancer, Ly: lymphoma, M: melanoma, My: myeloma, NAFLD: non-alcoholic fatty liver disease, NSCLC: non-small cell lung carcinoma, OC: ovarian cancer, PaC: pancreatic cancer, PC: prostate cancer, RCC: renal cell carcinoma, P: clinical trial phase, A: FDA-approved

## Supplementary Note 1 KEGG's documented cancer-related gene list (463).

ABL1, AGT, AGTR1, AKT1, AKT2, AKT3, ALK, APAF1, APPL1, AR, ARAF, ARNT, ARNT2, BAD, BAK1, BAX, BBC3, BCL2, BCL2L1, BCL2L11, BCR, BDKRB1, BDKRB2, BID, BIRC2, BIRC3, BIRC5, BIRC7, BMP2, BMP4, BRAF, CALM1, CALM2, CALM3, CALML3, CALML4, CALML5, CALML6, CAMK2A, CAMK2B, CAMK2D, CAMK2G, CASP3, CASP7, CASP8, CASP9, CBL, CCDC6, CCNA1, CCNA2, CCND1, CCND2, CCND3, CCNE1, CCNE2, CDC42, CDK2, CDK4, CDK6, CDKN1A, CDKN1B, CDKN2A, CDKN2B, CEBPA, CHUK, CKS1B, CKS2, COL4A1, COL4A2, COL4A3, COL4A4, COL4A5, COL4A6, CREBBP, CRK, CRKL, CSF1R, CSF2RA, CSF2RB, CSF3R, CTBP1, CTBP2, CTNNB1, CUL1, CUL2, CXCL12, CXCL8, CXCR4, CYCS, DAPK1, DAPK2, DAPK3, DCC, DDB2, DLL1, DLL3, DLL4, DVL1, DVL2, DVL3, E2F1, E2F2, E2F3, EDN1, EDNRA, EDNRB, EGF, EGFR, EGLN1, EGLN2, EGLN3, ELK1, ELOB, ELOC, EML4, EP300, EPAS1, EPO, EPOR, ERBB2, ESR1, ESR2, ETS1, F2, F2R, F2RL3, FADD, FAS, FASLG, FGF1, FGF10, FGF16, FGF17, FGF18, FGF19, FGF2, FGF20, FGF21, FGF22, FGF23, FGF3, FGF4, FGF5, FGF6, FGF7, FGF8, FGF9, FGFR1, FGFR2, FGFR3, FGFR4, FLT3, FLT3LG, FLT4, FN1, FOS, FOXO1, FRAT1, FRAT2, FZD1, FZD10, FZD2, FZD3, FZD4, FZD5, FZD6, FZD7, FZD8, FZD9, GADD45A, GADD45B, GADD45G, GLI1, GLI2, GLI3, GNA11, GNAQ, GRB2, GSK3B, GSTA1, GSTA2, GSTA3, GSTA4, GSTA5, GSTM1, GSTM2, GSTM3, GSTM4, GSTM5, GSTO1, GSTO2, GSTP1, GSTT2B, HDAC1, HDAC2, HES1, HES5, HEY1, HEY2, HEYL, HGF, HHIP, HIF1A, HMOX1, HRAS, HSP90AA1, HSP90AB1, HSP90B1, IFNA1, IFNA10, IFNA13, IFNA14, IFNA16, IFNA17, IFNA2, IFNA21, IFNA4, IFNA5, IFNA6, IFNA7, IFNA8, IFNAR1, IFNAR2, IFNG, IFNGR1, IFNGR2, IGF1, IGF1R, IGF2, IKBKB, IKBKG, IL12A, IL12B, IL12RB1, IL12RB2, IL13, IL13RA1, IL15, IL15RA, IL2, IL23A, IL23R, IL2RA, IL2RB, IL2RG, IL3, IL3RA, IL4, IL4R, IL5, IL5RA, IL6, IL6R, IL6ST, IL7, IL7R, ITGA2, ITGA2B, ITGA3, ITGA6, ITGAV, ITGB1, JAG1, JAG2, JAK1, JAK2, JAK3, JUN, JUP, KEAP1, KIF7, KIT, KITLG, KLK3, KNG1, KRAS, LAMA1, LAMA2, LAMA3, LAMA4, LAMA5, LAMB1, LAMB2, LAMB3, LAMB4, LAMC1, LAMC2, LAMC3, LEF1, LPAR1, LPAR2, LPAR3, LPAR4, LPAR5, LPAR6, LRP5, LRP6, MAP2K1, MAP2K2, MAPK1, MAPK10, MAPK3, MAPK8, MAPK9, MAX, MDM2, MECOM, MET, MGST1, MGST2, MGST3, MMP1, MMP2, MMP9, MTOR, MYC, NCOA1, NCOA3, NCOA4, NFE2L2, NFKB1, NFKB2, NFKBIA, NOS2, NOTCH1, NOTCH2, NOTCH3, NOTCH4, NQO1, NRAS, NTRK1, PAX8, PDGFA, PDGFB, PDGFRA, PDGFRB, PGF, PIK3CA, PIK3CB, PIK3CD, PIK3R1, PIK3R2, PIK3R3, PIM1, PIM2, PLCB1, PLCB2, PLCB3, PLCB4, PLCG1, PLCG2, PLD1, PLD2, PMAIP1, PML, POLK, PPARD, PPARG, PRKCA, PRKCB, PRKCG, PTCH1, PTCH2, PTGS2, PTK2, RAC1, RAC2, RAC3, RAF1, RALA, RALB, RALBP1, RALGDS, RARA, RASGRP1, RASGRP2, RASGRP3, RASGRP4, RASSF1, RASSF5, RB1, RBX1, RELA, RET, RHOA, RPS6KA5, RPS6KB1, RPS6KB2, RUNX1, RUNX1T1, RXRA, RXRB, RXRG, SHH, SKP1, SKP2, SLC2A1, SMAD2, SMAD3, SMAD4, SMO, SOS1, SOS2, SP1, SPI1, STAT1, STAT2, STAT3, STAT4, STAT5A, STAT5B, STAT6, STK4, SUFU, TCF7, TCF7L1, TCF7L2, TFG, TGFA, TGFB1, TGFB2, TGFB3, TGFBR1, TGFBR2, TP53, TPM3, TPR, TRAF1, TRAF2, TRAF3, TRAF4, TRAF5, TRAF6, TXNRD1, TXNRD2, TXNRD3, VEGFA, VEGFB, VEGFC, VEGFD, VHL, WNT1, WNT10A, WNT10B, WNT11, WNT16, WNT2, WNT2B, WNT3, WNT3A, WNT4, WNT5A, WNT5B, WNT6, WNT7A, WNT7B, WNT8A, WNT8B, WNT9A, WNT9B, XIAP, ZBTB16, ZBTB17

**Supplementary Table 1a Cancer-wise cell line counts: pretraining data**

| Cancer type | Count |
|---|---|
| Bile duct | 29 |
| Bladder | 14 |
| Bone | 28 |
| Brain | 49 |
| Breast | 40 |
| Cervical | 18 |
| Colorectal | 46 |
| Endometrial | 19 |
| Esophageal | 8 |
| Eye | 11 |
| Gastric | 25 |
| Head and Neck | 34 |
| Kidney | 19 |
| Leukemia | 104 |
| Liposarcoma | 10 |
| Liver | 7 |
| Lung | 105 |
| Lymphoma | 83 |
| Myeloma | 30 |
| Neuroblastoma | 26 |
| Ovarian | 34 |
| Pancreatic | 20 |
| Prostate | 8 |
| Rhabdoid | 15 |
| Sarcoma | 29 |
| Skin | 46 |
| Thyroid | 7 |
| Total | 864 |

Underlined cancer types indicate those which are plotted in **Figure 4** to examine embeddings.

**Supplementary Table 1b Cancer-wise cell line counts: novel cancer test set**

| Cancer type | Count |
|---|---|
| Bile Duct | 6 |
| Bone | 9 |
| Gallbladder | 1 |
| Gastric | 14 |
| Kidney | 14 |
| Neuroblastoma | 3 |
| Prostate | 2 |
| Rhabdoid | 4 |
| Sarcoma | 6 |
| Thyroid | 8 |
| Total | 67 |

**Supplementary Table 1c Cancer-wise cell line counts: training folds and trained-on cancer test set**

| Cancer type | Count | | | | | All folds | Testing |
|---|---|---|---|---|---|---|---|
| | Fold 1 | Fold 2 | Fold 3 | Fold 4 | Fold 5 | | |
| Bladder | 3 | 3 | 3 | 3 | 5 | 17 | 3 |
| Brain | 5 | 5 | 5 | 5 | 7 | 27 | 5 |
| Breast | 3 | 3 | 3 | 3 | 3 | 15 | 3 |
| Colorectal | 4 | 4 | 4 | 4 | 4 | 20 | 4 |
| Endometrial | 3 | 3 | 3 | 3 | 5 | 17 | 3 |
| Esophageal | 4 | 4 | 4 | 4 | 2 | 18 | 3 |
| Head and Neck | 3 | 3 | 3 | 3 | 5 | 17 | 3 |
| Liver | 3 | 3 | 3 | 3 | 1 | 13 | 2 |
| Lung | 15 | 15 | 15 | 15 | 16 | 76 | 13 |
| Ovarian | 5 | 5 | 5 | 5 | 5 | 25 | 4 |
| Pancreatic | 5 | 5 | 5 | 5 | 5 | 25 | 4 |
| Skin | 6 | 6 | 6 | 6 | 6 | 30 | 5 |
| Total | 59 | 59 | 59 | 59 | 64 | 300 | 52 |

**Supplementary Table 2 Hyperparameter options: SiamCDR's drug and cell line encoders (*Enc*) trained by siamese neural networks (SNN).**

| Hyperparameter | Options tested |
|---|---|
| Number of hidden layers | *1*, **2** |
| Number of units per layer | <u>16</u>, 32, ***64*** |
| Activation function | ***ReLU***, sigmoid |
| Dropout rate | 0.0, ***0.1***, 0.3 |
| Learning rate | <u>0.01</u>, ***0.001***, 0.0001 |
| Decay rate | <u>0.99</u> |
| Decay steps | <u>1024</u> |
| Patience | <u>10</u> |
| Minimum delta | <u>0.0001</u> |
| Minibatch size | <u>512</u> |
| Maximum number of epochs | <u>1000</u> |

For hyperparameters with more than one option, an <u>underlined value</u>, an *italicized value*, or a **bold** value denotes the hyperparameter option with best performing <u>$D_{Embed}$ with $C_{Raw}$</u> or *$D_{Embed}$ with $C_{Embed}$*, or best performing **$D_{Raw}$ with $C_{Embed}$**, respectively.

**Supplementary Table 3a Hyperparameter options: Random forest (RF) classifier.**

| Hyperparameter | Options tested |
|---|---|
| Criterion | <u>gini</u>, entropy |
| Number of estimators | 10, 25, 50, <u>100</u> |
| Minimum samples to split | 5, 10, <u>20</u>, 25 |

<u>Underlined values</u> denote the hyperparameter option with best performance.

**Supplementary Table 3b Hyperparameters options: DNN classifier.**

| Hyperparameter | Options tested |
|---|---|
| Dimensions of hidden layers | 64-32-16; 64-32-8; 64-16-8; 32-16-8; 64-64-64; 32-32-32; 16-16-16; <u>64-64</u>; 32-32; 16-16; 64-32; 64-16; 32-16; 36; 32; 16 |
| Activation function | <u>ReLU</u>, sigmoid |
| Dropout rate | 0.0, <u>0.1</u>, 0.3 |
| Learning rate | <u>0.01</u>, 0.001 |
| Decay rate | <u>0.99</u> |
| Decay steps | 50, <u>500</u> |
| Patience | <u>10</u> |
| Minimum delta | <u>0.0001</u> |
| Minibatch size | <u>256</u> |
| Maximum epochs | <u>1000</u> |

<u>Underlined values</u> denote the hyperparameter option with best performance.

**Supplementary Table 4a** $P_{\text{cell}}@k$ **for trained-on cancers.**

| Drug | Cell line | Classifier | $P_{\text{cell}}@k$ | | | | | |
|------|-----------|------------|------|------|------|------|------|------|
| | | | 1 | 2 | 3 | 4 | 5 | 10 |
| f | g | | *0.8235* | *0.8824* | *0.8824* | *0.8382* | *0.7843* | *0.7000* |
| e_d | | Logistic | 0.7451 | 0.6471 | 0.5621 | 0.4902 | 0.4588 | 0.3667 |
| f | e_c | (LR) | 0.9412 | 0.9112 | 0.8693 | 0.8593 | 0.8275 | 0.8077 |
| e_d | | | 0.9412 | 0.9020 | 0.8497 | 0.8480 | 0.8275 | 0.8128 |
| f | g | | *0.9412* | *0.8922* | *0.9020* | *0.8676* | *0.8392* | *0.8385* |
| e_d | | Random | 0.8039 | 0.8529 | 0.8366 | 0.8333 | 0.8118 | 0.7718 |
| f | e_c | Forest (RF) | 0.9412 | 0.9020 | 0.8824 | 0.8676 | **0.8588** | 0.8103 |
| e_d | | | 0.9608 | **0.9412** | **0.9085** | **0.8775** | 0.8549 | 0.8026 |
| f | g | | *0.9412* | *0.9020* | *0.8889* | *0.8578* | *0.8471* | *0.8308* |
| e_d | | DNN | 0.8235 | 0.6471 | 0.5686 | 0.5086 | 0.4784 | 0.4282 |
| f | e_c | | **0.9804** | **0.9412** | 0.8654 | 0.8529 | 0.8235 | **0.8436** |
| e_d | | | **0.9804** | 0.9216 | 0.9020 | 0.8529 | 0.8471 | 0.8385 |

For model architecture, underlined parameters denote highest performing model for that classifier. **Bolded values** denote top performance for that metric across all model variations; underlined values denote top performance for that metric across all variations of a given end classifier (i.e. LR, RF, etc); and *italicized values* denote a classifier's vanilla performance without any feature embedding.

**Supplementary Table 4b** $P_{\text{cell}}@k$ **for novel cancers.**

| Drug | Cell line | Classifier | $P_{\text{cell}}@k$ | | | | | |
|------|-----------|------------|------|------|------|------|------|------|
| | | | 1 | 2 | 3 | 4 | 5 | 10 |
| f | g | | *0.8923* | *0.9154* | *0.8615* | *0.8423* | *0.8000* | *0.7157* |
| e_d | | Logistic | 0.7385 | 0.6615 | 0.6000 | 0.5231 | 0.4954 | 0.4294 |
| f | e_c | (LR) | 0.9538 | 0.9077 | **0.9077** | 0.8731 | 0.8585 | 0.8020 |
| e_d | | | 0.9538 | 0.9077 | 0.9026 | 0.8731 | 0.8615 | 0.8078 |
| f | g | | *0.9538* | *0.9154* | *0.8821* | *0.8654* | *0.8554* | *0.8235* |
| e_d | | Random | 0.8308 | 0.8538 | 0.8462 | 0.8385 | 0.8154 | 0.7843 |
| f | e_c | Forest (RF) | **0.9846** | 0.9385 | 0.8718 | 0.8423 | 0.8185 | 0.7922 |
| e_d | | | 0.9538 | **0.9538** | 0.8923 | 0.8500 | 0.8277 | 0.7941 |
| f | g | | *0.9692* | *0.9231* | *0.9026* | *0.8885* | *0.8585* | *0.8157* |
| e_d | | DNN | 0.7692 | 0.6538 | 0.6103 | 0.5385 | 0.5077 | 0.4765 |
| f | e_c | | **0.9846** | 0.9462 | 0.9026 | 0.8692 | 0.8615 | 0.8275 |
| e_d | | | 0.9538 | 0.9154 | 0.8923 | 0.8769 | **0.8738** | 0.8216 |

**Bolded values** denote top performance for that metric across all model variations; underlined values denote top performance for that metric across all variations of a given end classifier (i.e. LM, RF, etc); and *italicized values* denote a classifier's vanilla performance without any feature embedding.

**Supplementary Table 5a** $P_{\text{cancer}}@k$ **for trained-on cancers**

| Model Architecture | | | $P_{\text{cancer}}@k$ | | | | |
|---|---|---|---|---|---|---|---|
| Drug | Cell line | Classifier | 1 | 2 | 3 | 4 | 5 |
| **f** | **g** | | *0.8419* | *0.8862* | <u>*0.8912*</u> | *0.8526* | *0.8019* |
| **e_d** | | Logistic | 0.7311 | 0.6386 | 0.5554 | 0.4913 | 0.4564 |
| <u>**f**</u> | <u>**e_c**</u> | (LR) | <u>0.9450</u> | <u>0.9265</u> | 0.8824 | <u>0.8564</u> | <u>0.8280</u> |
| **e_d** | | | <u>0.9450</u> | 0.9161 | 0.8569 | 0.8505 | <u>0.8280</u> |
| **f** | **g** | | *0.9450* | *0.9043* | *0.9109* | *0.8838* | *0.8484* |
| **e_d** | | Random | 0.8363 | 0.8612 | 0.8540 | 0.8420 | 0.8137 |
| **f** | | Forest (RF) | 0.9663 | 0.9288 | 0.8916 | 0.8713 | <u>**0.8606**</u> |
| <u>**e_d**</u> | <u>**e_c**</u> | | <u>0.9728</u> | **0.9591** | **0.9171** | <u>0.8850</u> | 0.8571 |
| **f** | **g** | | *0.9450* | *0.9160* | *0.8933* | *0.8600* | *0.8453* |
| **e_d** | | DNN | 0.8283 | 0.6427 | 0.5647 | 0.5016 | 0.4814 |
| <u>**f**</u> | <u>**e_c**</u> | | **0.9792** | <u>0.9378</u> | 0.9040 | <u>0.8663</u> | 0.8327 |
| **e_d** | | | **0.9792** | 0.9297 | 0.9109 | 0.8593 | 0.8517 |

**Bolded values** denote top performance for that metric across all model variations; <u>underlined values</u> denote top performance for that metric across all variations of a given end classifier (i.e. LM, RF, etc); and *italicized values* denote a classifier's vanilla performance without any feature embedding.

**Supplementary Table 5b** $P_{\text{cancer}}@k$ **for novel cancers.**

| Model Architecture | | | $P_{\text{cancer}}@k$ | | | | |
|---|---|---|---|---|---|---|---|
| Drug | Cell line | Classifier | 1 | 2 | 3 | 4 | 5 |
| **f** | **g** | | *0.9290* | <u>*0.9361*</u> | *0.8597* | *0.8204* | *0.7886* |
| **e_d** | | Logistic | 0.6638 | 0.6544 | 0.5808 | 0.5053 | 0.4730 |
| <u>**f**</u> | <u>**e_c**</u> | (LR) | <u>0.9646</u> | 0.9222 | **0.9289** | <u>0.8861</u> | 0.8485 |
| **e_d** | | | <u>0.9646</u> | 0.9222 | 0.9263 | <u>0.8861</u> | <u>0.8552</u> |
| **f** | **g** | | *0.9721* | *0.9460* | <u>*0.8904*</u> | *0.8605* | <u>*0.8609*</u> |
| **e_d** | | Random | 0.8106 | 0.8524 | 0.8552 | 0.8284 | 0.7919 |
| **f** | | Forest (RF) | **0.9923** | 0.9505 | 0.8544 | 0.8315 | 0.8068 |
| <u>**e_d**</u> | <u>**e_c**</u> | | 0.9769 | **0.9641** | 0.8820 | <u>0.8856</u> | 0.8187 |
| <u>**f**</u> | <u>**g**</u> | | *0.9846* | *0.9347* | <u>*0.9239*</u> | **_0.8953_** | <u>**_0.8663_**</u> |
| **e_d** | | DNN | 0.7710 | 0.6461 | 0.6224 | 0.5390 | 0.5130 |
| **f** | **e_c** | | **0.9923** | <u>0.9542</u> | 0.9064 | 0.8573 | 0.8290 |
| **e_d** | | | 0.9769 | 0.9285 | 0.8764 | 0.8706 | 0.8582 |

**Bolded values** denote top performance for that metric across all model variations; <u>underlined values</u> denote top performance for that metric across all variations of a given end classifier (i.e. LM, RF, etc); and *italicized values* denote a classifier's vanilla performance without any feature embedding.

**Supplementary Table 6a DeepDSC's $P_{cancer}@k$ for trained-on cancers.**

| Cancer type | $P_{cancer}@1$ | $P_{cancer}@2$ | $P_{cancer}@3$ | $P_{cancer}@4$ | $P_{cancer}@5$ |
|---|---|---|---|---|---|
| Bladder | 1.0000 | 0.8333 | 0.8889 | 0.8333 | 0.7333 |
| Brain | 0.6667 | 0.8333 | 0.5556 | 0.5833 | 0.6000 |
| Breast | 1.0000 | 0.8333 | 0.5556 | 0.6667 | 0.5333 |
| Colorectal | 1.0000 | 0.8750 | 0.7500 | 0.6250 | 0.5500 |
| Endometrial | 1.0000 | 0.8333 | 0.7778 | 0.7500 | 0.6667 |
| Esophageal | 1.0000 | 1.0000 | 0.8333 | 0.7500 | 0.7000 |
| Head and Neck | 1.0000 | 1.0000 | 0.6667 | 0.5833 | 0.6000 |
| Liver | 1.0000 | 1.0000 | 0.7556 | 0.7500 | 0.6000 |
| Lung | 0.9333 | 0.9000 | 0.6667 | 0.7333 | 0.6667 |
| Ovarian | 0.6000 | 0.8000 | 0.6667 | 0.7000 | 0.6400 |
| Pancreatic | 0.8000 | 0.8000 | 0.6667 | 0.7000 | 0.6800 |
| Skin | 1.0000 | 0.8333 | 0.6667 | 0.5833 | 0.5333 |
| Overall | 0.9167 | 0.8785 | 0.7227 | 0.6882 | 0.6253 |

**Supplementary Table 6b SiamCDR$_{RF}$'s $P_{cancer}@k$ for trained-on cancers.**

| Cancer type | $P_{cancer}@1$ | $P_{cancer}@2$ | $P_{cancer}@3$ | $P_{cancer}@4$ | $P_{cancer}@5$ |
|---|---|---|---|---|---|
| Bladder | 1.0000 | 1.0000 | 1.0000 | 0.9167 | 0.9333 |
| Brain | 1.0000 | 0.8750 | 0.8333 | 0.8125 | 0.7500 |
| Breast | 1.0000 | 1.0000 | 0.7778 | 0.7500 | 0.8000 |
| Colorectal | 1.0000 | 1.0000 | 1.0000 | 0.8750 | 0.9000 |
| Endometrial | 1.0000 | 1.0000 | 1.0000 | 0.9167 | 0.8667 |
| Esophageal | 1.0000 | 0.8333 | 0.7778 | 0.7500 | 0.8000 |
| Head and Neck | 1.0000 | 1.0000 | 1.0000 | 1.0000 | 0.9333 |
| Liver | 1.0000 | 1.0000 | 1.0000 | 1.0000 | 0.9000 |
| Lung | 0.9231 | 0.9231 | 0.8974 | 0.8846 | 0.8615 |
| Ovarian | 1.0000 | 1.0000 | 1.0000 | 0.8750 | 0.8615 |
| Pancreatic | 0.7500 | 0.8750 | 0.8333 | 0.8125 | 0.8000 |
| Skin | 1.0000 | 1.0000 | 0.8667 | 0.9000 | 0.8000 |
| Overall | 0.9728 | 0.9589 | 0.9155 | 0.8744 | 0.8454 |

**Supplementary Table 6c SiamCDR$_{LR}$'s $P_{cancer}@k$ for trained-on cancers.**

| Cancer type | $P_{cancer}@1$ | $P_{cancer}@2$ | $P_{cancer}@3$ | $P_{cancer}@4$ | $P_{cancer}@5$ |
|---|---|---|---|---|---|
| Bladder | 1.0000 | 1.0000 | 1.0000 | 1.0000 | 1.0000 |
| Brain | 1.0000 | 0.8750 | 0.8333 | 0.7500 | 0.7500 |
| Breast | 1.0000 | 1.0000 | 0.7778 | 0.8333 | 0.8000 |
| Colorectal | 1.0000 | 1.0000 | 0.9167 | 0.9375 | 0.9000 |
| Endometrial | 1.0000 | 1.0000 | 0.8889 | 0.9167 | 0.8667 |
| Esophageal | 0.6667 | 0.8333 | 0.8889 | 0.7500 | 0.8000 |
| Head and Neck | 1.0000 | 1.0000 | 0.8889 | 0.8333 | 0.8000 |
| Liver | 1.0000 | 1.0000 | 1.0000 | 1.0000 | 1.0000 |
| Lung | 0.9231 | 0.8846 | 0.8718 | 0.9038 | 0.8462 |
| Ovarian | 1.0000 | 0.8750 | 0.8333 | 0.8125 | 0.8500 |
| Pancreatic | 0.7500 | 0.7500 | 0.8333 | 0.7500 | 0.7500 |
| Skin | 1.0000 | 0.9000 | 0.8000 | 0.8500 | 0.8000 |
| Overall | 0.9450 | 0.9265 | 0.8777 | 0.8614 | 0.8469 |

**Supplementary Table 6d SiamCDR$_{DNN}$'s $P_{cancer}@k$ for trained-on cancers.**

| Cancer type | $P_{cancer}@1$ | $P_{cancer}@2$ | $P_{cancer}@3$ | $P_{cancer}@4$ | $P_{cancer}@5$ |
|---|---|---|---|---|---|
| Bladder | 1.0000 | 1.0000 | 1.0000 | 1.0000 | 0.9333 |
| Brain | 1.0000 | 0.8750 | 0.8333 | 0.8125 | 0.7500 |
| Breast | 1.0000 | 0.8333 | 0.8889 | 0.8333 | 0.8667 |
| Colorectal | 1.0000 | 1.0000 | 0.9167 | 0.9375 | 0.8500 |
| Endometrial | 1.0000 | 1.0000 | 0.8889 | 0.9167 | 0.8667 |
| Esophageal | 1.0000 | 0.8333 | 0.8889 | 0.8333 | 0.7333 |
| Head and Neck | 1.0000 | 1.0000 | 1.0000 | 0.9167 | 0.9333 |
| Liver | 1.0000 | 1.0000 | 1.0000 | 1.0000 | 1.0000 |
| Lung | 0.9231 | 0.8462 | 0.8974 | 0.8654 | 0.8769 |
| Ovarian | 1.0000 | 1.0000 | 0.9167 | 0.9375 | 0.8000 |
| Pancreatic | 0.7500 | 0.7500 | 0.8333 | 0.8125 | 0.7500 |
| Skin | 1.0000 | 1.0000 | 0.8667 | 0.8500 | 0.8000 |
| Overall | 0.9728 | 0.9282 | 0.9109 | 0.8929 | 0.8467 |

**Supplementary Table 7a DeepDSC's $P_{\text{cancer}}@k$ for novel cancer types.**

| Cancer type | $P_{\text{cancer}}@1$ | $P_{\text{cancer}}@2$ | $P_{\text{cancer}}@3$ | $P_{\text{cancer}}@4$ | $P_{\text{cancer}}@5$ |
|---|---|---|---|---|---|
| Bile | 0.6000 | 0.7000 | 0.6667 | 0.5550 | 0.4400 |
| Bone | 1.0000 | 0.8889 | 0.7407 | 0.6667 | 0.6444 |
| Gallbladder | 1.0000 | 1.0000 | 0.6667 | 0.5000 | 0.4000 |
| Gastric | 1.0000 | 0.9286 | 0.8095 | 0.7857 | 0.7429 |
| Kidney | 0.6154 | 0.6583 | 0.6923 | 0.6154 | 0.5846 |
| Neuroblastoma | 1.0000 | 1.0000 | 0.8889 | 0.9167 | 0.9333 |
| Prostate | 1.0000 | 1.0000 | 0.8333 | 0.8750 | 0.7000 |
| Rhabdoid | 0.7500 | 0.7500 | 0.6667 | 0.6875 | 0.6500 |
| Sarcoma | 0.8333 | 0.8333 | 0.7778 | 0.7083 | 0.6000 |
| Thyroid | 0.7500 | 0.8125 | 0.6667 | 0.7188 | 0.6750 |
| Overall | 0.8549 | 0.8567 | 0.7409 | 0.7024 | 0.6370 |

**Supplementary Table 7b SiamCDR$_{\text{RF}}$'s $P_{\text{cancer}}@k$ for novel cancer types.**

| Cancer type | $P_{\text{cancer}}@1$ | $P_{\text{cancer}}@2$ | $P_{\text{cancer}}@3$ | $P_{\text{cancer}}@4$ | $P_{\text{cancer}}@5$ |
|---|---|---|---|---|---|
| Bile | 1.0000 | 0.9000 | 0.9333 | 0.9000 | 0.8800 |
| Bone | 1.0000 | 1.0000 | 0.9630 | 0.9167 | 0.8889 |
| Gallbladder | 1.0000 | 1.0000 | 1.0000 | 0.7500 | 0.6000 |
| Gastric | 1.0000 | 0.9643 | 0.9524 | 0.9286 | 0.9286 |
| Kidney | 0.7692 | 0.7692 | 0.7179 | 0.6923 | 0.6769 |
| Neuroblastoma | 1.0000 | 0.8333 | 0.7778 | 0.6667 | 0.7333 |
| Prostate | 1.0000 | 1.0000 | 1.0000 | 1.0000 | 1.0000 |
| Rhabdoid | 1.0000 | 1.0000 | 0.9167 | 0.8750 | 0.9000 |
| Sarcoma | 0.8333 | 0.9167 | 0.9444 | 0.8750 | 0.8667 |
| Thyroid | 1.0000 | 0.9375 | 0.9583 | 0.9062 | 0.8500 |
| Overall | 0.9603 | 0.9321 | 0.9164 | 0.8510 | 0.8324 |

**Supplementary Table 7c SiamCDR$_{\text{LR}}$'s $P_{\text{cancer}}@k$ for novel cancer types.**

| Cancer type | $P_{\text{cancer}}@1$ | $P_{\text{cancer}}@2$ | $P_{\text{cancer}}@3$ | $P_{\text{cancer}}@4$ | $P_{\text{cancer}}@5$ |
|---|---|---|---|---|---|
| Bile Duct | 0.8000 | 0.8000 | 0.8667 | 0.8500 | 0.8000 |
| Bone | 1.0000 | 0.9444 | 0.8889 | 0.8611 | 0.8667 |
| Gallbladder | 1.0000 | 1.0000 | 1.0000 | 0.7500 | 0.6000 |
| Gastric | 1.0000 | 1.0000 | 1.0000 | 0.9643 | 0.9429 |
| Kidney | 0.8462 | 0.7692 | 0.7436 | 0.7308 | 0.7077 |
| Neuroblastoma | 1.0000 | 0.8333 | 0.8889 | 0.9167 | 0.9333 |
| Prostate | 1.0000 | 1.0000 | 1.0000 | 1.0000 | 0.9000 |
| Rhabdoid | 1.0000 | 1.0000 | 1.0000 | 0.9375 | 0.9500 |
| Sarcoma | 1.0000 | 1.0000 | 1.0000 | 1.0000 | 0.9333 |
| Thyroid | 1.0000 | 0.9375 | 0.9167 | 0.9062 | 0.9000 |
| Overall | 0.9646 | 0.9285 | 0.9305 | 0.8917 | 0.8534 |

**Supplementary Table 7d SiamCDR$_{\text{DNN}}$'s $P_{\text{cancer}}@k$ for novel cancer types.**

| Cancer type | $P_{\text{cancer}}@1$ | $P_{\text{cancer}}@2$ | $P_{\text{cancer}}@3$ | $P_{\text{cancer}}@4$ | $P_{\text{cancer}}@5$ |
|---|---|---|---|---|---|
| Bile Duct | 1.0000 | 0.9000 | 0.8667 | 0.8500 | 0.8800 |
| Bone | 1.0000 | 0.9444 | 0.9259 | 0.8889 | 0.9111 |
| Gallbladder | 1.0000 | 1.0000 | 1.0000 | 0.7500 | 0.8000 |
| Gastric | 1.0000 | 1.0000 | 0.9524 | 0.9464 | 0.9142 |
| Kidney | 0.9231 | 0.8077 | 0.7949 | 0.7500 | 0.7231 |
| Neuroblastoma | 1.0000 | 1.0000 | 1.0000 | 0.8333 | 0.8000 |
| Prostate | 1.0000 | 1.0000 | 1.0000 | 1.0000 | 0.9000 |
| Rhabdoid | 1.0000 | 1.0000 | 0.9167 | 0.9375 | 0.9000 |
| Sarcoma | 0.8333 | 0.9167 | 0.9444 | 0.9167 | 0.8667 |
| Thyroid | 1.0000 | 1.0000 | 0.9583 | 0.9375 | 0.8500 |
| Overall | 0.9756 | 0.9569 | 0.9359 | 0.8810 | 0.8545 |

**Supplementary Table 8 Mechanisms of action (MOA) with at least 10 drugs in pretraining data.**

| MOA | Count |
|---|---|
| EGFR inhibitor | 32 |
| HDAC inhibitor | 25 |
| Glucocorticoid receptor agonist | 24 |
| Topoisomerase inhibitor | 23 |
| Tubulin polymerization inhibitor | 21 |
| Aurora kinase inhibitor | 18 |
| Histamine receptor antagonist | 18 |
| MEK inhibitor | 18 |
| Adrenergic receptor antagonist | 17 |
| PI3K inhibitor | 16 |
| Protein synthesis inhibitor | 16 |
| mTOR inhibitor | 16 |
| CDK inhibitor | 16 |
| Cyclooxygenase inhibitor | 15 |
| Adrenergic receptor agonist | 14 |
| Dopamine receptor antagonist | 13 |
| Glutamate receptor antagonist | 13 |
| HSP inhibitor | 13 |
| Sodium channel blocker | 11 |
| Benzodiazepine receptor agonist | 11 |
| Phosphodiesterase inhibitor | 10 |
| DNA inhibitor | 10 |
| Total | 370 |

**Supplementary Table 9 Cancer type ranking of FDA indicated drugs.**
Mean rank of FDA indicated drugs per each cell line is reported for both DeepDSC and our best SiamCDR model variation.
Additionally, for each model the best rank is reported with the drug denoted in parenthesis. Due to length of this table, it is available
at the following link: https://github.com/ninglab/SiamCDR/blob/main/res/SupplementaryTable9.xlsx

**Supplementary Table 10 Significant Spearman correlations: SiamCDR_RF's docetaxel priority ranking and gene expression of breast cancer cells.**

| Gene name | Correlation | *p*-value |
|-----------|-------------|-----------|
| PLD2 | -0.5037 | 0.0006 |
| CALM1 | -0.4918 | 0.0008 |
| LRP5 | -0.4898 | 0.0009 |
| RASGRP2 | -0.4886 | 0.0009 |
| WNT9A | -0.4854 | 0.001 |
| VEGFB | -0.4611 | 0.0019 |
| TRAF2 | -0.4609 | 0.0019 |
| CCND3 | -0.4578 | 0.002 |
| PLCB3 | -0.4553 | 0.0022 |
| CAMK2G | -0.446 | 0.0027 |
| NCOA1 | -0.4418 | 0.003 |
| BAD | -0.4394 | 0.0032 |
| CSF2RB | -0.4257 | 0.0044 |
| WNT3A | -0.4202 | 0.005 |
| AKT1 | -0.4193 | 0.0051 |
| TRAF4 | -0.4182 | 0.0053 |
| FGFR2 | -0.4151 | 0.0056 |
| CSF3R | -0.4068 | 0.0068 |
| RALB | -0.4063 | 0.0069 |
| TGFBR1 | -0.3963 | 0.0085 |
| RPS6KB2 | -0.3928 | 0.0092 |
| WNT7B | -0.3855 | 0.0107 |
| TCF7L1 | -0.3805 | 0.0118 |
| IL2RB | -0.3788 | 0.0122 |
| LAMA5 | -0.3775 | 0.0126 |
| RELA | -0.3755 | 0.0131 |
| PLCB4 | -0.3732 | 0.0137 |
| FGF22 | -0.3708 | 0.0144 |
| PDGFB | -0.3694 | 0.0148 |
| GNA11 | -0.3682 | 0.0151 |
| HES5 | -0.3659 | 0.0158 |
| MAP2K2 | -0.3646 | 0.0162 |
| BAK1 | -0.3578 | 0.0185 |
| IKBKG | -0.3555 | 0.0193 |
| EPOR | -0.3526 | 0.0204 |
| BDKRB2 | -0.3518 | 0.0207 |
| LPAR1 | 0.372 | 0.014 |
| LPAR6 | 0.3788 | 0.0123 |
| LPAR3 | 0.3872 | 0.0103 |
| F2R | 0.3904 | 0.0096 |

**Supplementary Table 11 Data, hardware, and major software used.**

| Data set | Source (DOI) | Version |
|---|---|---|
| PRISM repurposing secondary screen | https://doi.org/10.1038/s43018-019-0018-6 | PRISM Repurposing 19Q4 |
| Cell line gene expression and cancer types | https://doi.org/10.1038/s41586-019-1186-3 | DepMap Public 22Q2 |
| **Hardware** | **Source** | **Version** |
| CPU | Ohio Supercomputer Center | Intel Xeon 8268s Cascade Lakes |
| GPU | Ohio Supercomputer Center | NVIDIA Volta V100 |
| **Software** | **Version / DOI** | |
| Python | 3.8.12 | |
| Tensorflow | 2.8.0 | |
| Scikit-learn | 1.0.2 | |
| DeepDSC | 10.1109/TCBB.2019.2919581 | |